\documentclass[twoside]{article}

\usepackage[arxiv]{aistats2025}
\usepackage{amsmath,amsfonts,bm}

\def\eqref#1{equation~\ref{#1}}

\def\1{\bm{1}}

\DeclareMathAlphabet{\mathsfit}{\encodingdefault}{\sfdefault}{m}{sl}
\SetMathAlphabet{\mathsfit}{bold}{\encodingdefault}{\sfdefault}{bx}{n}

\DeclareMathOperator*{\argmax}{arg\,max}
\DeclareMathOperator*{\argmin}{arg\,min}

\usepackage{hyperref}
\usepackage{url}
\usepackage{csquotes}
\usepackage[style=authoryear-comp,natbib=true,maxcitenames=1,maxbibnames=999,uniquename=init,uniquelist=minyear]{biblatex}
\defbibheading{subbibliography}[\refname]{\subsubsection*{#1}}
\defbibfilter{appendixOnlyFilter}{
  segment=1 
  and not segment=0 
}
\usepackage{bookmark}
\usepackage{algorithmic}
\usepackage{algorithm}
\usepackage{graphicx,wrapfig}
\usepackage{subcaption}
\usepackage{enumitem}
\usepackage{mdframed}
\usepackage{amssymb}
\usepackage{standalone}
\usepackage{booktabs}
\usepackage{appendix}
\usepackage[disable]{todonotes}
\usepackage[english]{babel}

\addto\extrasenglish{%
}
\addto\extrasenglish{%
}
\addto\extrasenglish{%
}
\addto\extrasenglish{%
    
}
\addto\extrasenglish{%
    
}

\graphicspath{ {./images/} }

\renewcommand{\eqref}[1]{(\autoref{#1})}
\newcommand{\qed}{\tag*{$\blacksquare$}}

\usepackage{etoolbox}
\makeatletter
\patchcmd{\hyper@makecurrent}{%
    \ifx\Hy@param\Hy@chapterstring
        \let\Hy@param\Hy@chapapp
    \fi
}{%
    \iftoggle{inappendix}{
        \@checkappendixparam{chapter}%
        \@checkappendixparam{section}%
        \@checkappendixparam{subsection}%
        \@checkappendixparam{subsubsection}%
        \@checkappendixparam{paragraph}%
        \@checkappendixparam{subparagraph}%
    }{}%
}{}{\errmessage{failed to patch}}

\newcommand*{\@checkappendixparam}[1]{%
    \def\@checkappendixparamtmp{#1}%
    \ifx\Hy@param\@checkappendixparamtmp
        \let\Hy@param\Hy@appendixstring
    \fi
}
\makeatletter

\newtoggle{inappendix}
\togglefalse{inappendix}

\apptocmd{\appendix}{\toggletrue{inappendix}}{}{\errmessage{failed to patch}}
\apptocmd{\subappendices}{\toggletrue{inappendix}}{}{\errmessage{failed to patch}}

\begin{document}
%

%

\twocolumn[
    \aistatstitle{Dynamic Information Sub-Selection for Decision Support}
    \aistatsauthor{ Hung-Tien Huang,\  Maxwell Lennon,\  Shreyas Bhat Brahmavar,  Sean Sylvia,\ \&\  Junier B. Oliva }
    \aistatsaddress{
        University of North Carolina at Chapel Hill\\
        North Carolina, USA
    }
]

\begin{abstract}
    We introduce Dynamic Information Sub-Selection (DISS), a novel framework of AI assistance  designed to enhance the performance of black-box decision-makers by tailoring their information processing on a per-instance basis. Blackbox decision-makers (e.g., humans or real-time systems) often face challenges in processing all possible information at hand (e.g., due to cognitive biases or resource constraints), which can degrade decision efficacy. DISS addresses these challenges through policies that dynamically select the most effective features and options to forward to the black-box decision-maker for prediction. We develop a scalable frequentist data acquisition strategy and a decision-maker mimicking technique for enhanced budget efficiency. We explore several impactful applications of DISS, including biased decision-maker support, expert assignment optimization, large language model decision support, and interpretability. Empirical validation of our proposed DISS methodology shows superior performance to state-of-the-art methods across various applications.
\end{abstract}

\section{Introduction}
Typical machine learning (ML) approaches process all instance information consistently.
However, computational costs, biases, and other limitations of decision-makers often make it so that they are better served by a per-instance \emph{dynamic}
adjustment of what features are processed (and how to process them) when making predictions. For example, human decision-makers may actually do worse with more information on an instance due to cognitive overload \citep{iskander_burnout_2018}, thereby necessitating a concise summary of the most relevant features per instance \citep{morr-2018}.
Moreover, ML on edge devices such as microcontrollers may be subject to strict time and memory constraints \citep{plas-2018}, making it so that it is only possible to infer using a subset of features.
In this work, we introduce \emph{a new class of AI assistance} for black-box decision makers (BDMs) based on dynamic information sub-selection (DISS) that determines what pieces of information are effective for a BDM in a per-instance-personalized fashion; we develop novel methods to learn policies that tune what information to utilize when making decisions with BDMs.
\begin{figure}[t!]
      \centering
      \includegraphics[width=0.75\linewidth]{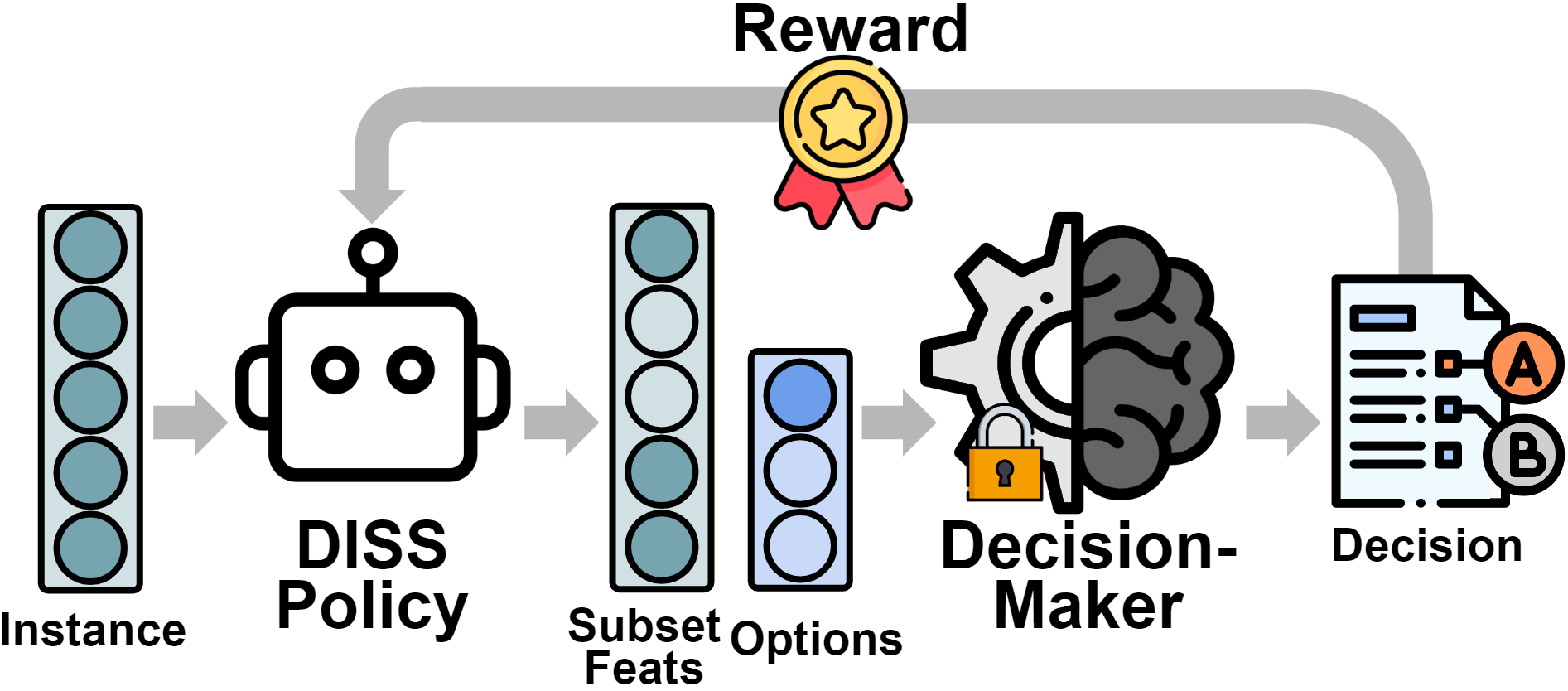}
      \caption{Given an instance, the DISS policy selects a subset of features (and options) to forward to a decision-maker to render its decisions. A reward that evaluates the quality of the decision maker's decision is used as feedback on the DISS policy.}
      \label{fig:graphical_abstract}
\end{figure}

DISS is a versatile framework that can be applied to a variety of real-world applications such as follows:

\textbf{Human Decision Support}\quad In many real-world applications (e.g., in law, human-resources, healthcare, etc.) humans must be ultimately responsible for decisions due to ethical, legal, or other reasons \citep{plas-2018,niel-2023}. However, human cognition has many well-studied limitations and biases \citep{grif-2020,busc-2011,caru-2020}, which may limit efficacy dependent on the presented information. Moreover, unlike statistical models, humans may process equivalent pieces of information differently depending on presentation \citep{lindsay2013human}. In these scenarios, DISS may be utilized to curate what and how information
is displayed to decision-makers.

\textbf{Expert Assignment}\quad One often has multiple decision-makers available, with each having their own set of expertise and decision-making processes, and this statement holds for both human experts \citep{mose-2009,moha-2001} and ML or statistical models \citep{zhou-2012}.
Below, we show how DISS may be used to not only select which decision-maker to employ in a per-instance fashion, but also customize the information presented to the selected expert, thereby maximizing the efficacy of decisions made by the expert.

\textbf{LLM Decision Support}\quad Recent research in Large Language Model (LLM) research has shown massive potential for applications in a variety of domains \citep{zhang2024llmmastermindsurveystrategic}. Yet, recent works \citep{liu-etal-2024-lost, Zhao2021CalibrateBU} have shown these models are sensitive to both prompting styles and content. DISS has the capability to personalize prompting style and content on a per-instance basis to achieve the best possible decisions with existing black-box LLMs such as GPT-4, LlaMA, Qwen2.5B \citep{zhao2024surveylargelanguagemodels}.

\textbf{Interpretability}\quad The set of most relevant features often varies from instance to instance. Prior works in interpretable ML have leveraged this observation for a better understanding of data and models \citep{saar-2021}. For example, existing methods often utilize proxy measurements such as gradients \citep{wang-2024} or model approximations \citep{lakk-2017} to estimate feature importance for each instance. DISS presents a novel, complementary approach for presenting feature importance by directly producing a dynamic subset of features solely responsible for the model's output whilst producing efficacious decisions. 

A core principle of this work is a data-driven approach that directly considers the efficacy of the BDM when given various potential subsets of features (and other options) for prediction on instances. Therefore, we pose DISS as a reward maximization problem: given an instance, the DISS policy selects information and options, passes its selection to the BDM, and obtains a reward according to the efficacy of the decision made by BDM (e.g., based on accuracy, confidence, time, etc.). See \autoref{fig:graphical_abstract} for an illustration. Therefore, obtaining decision-making observations directly from the BDM is vital to tuning the DISS policy. However, in many applications,
there are limitations to how many decisions one can query from the BDM during training due to restrictions in computational resources, time, monetary incentives, etc. Hence, this paper develops an acquisition strategy to effectively learn a policy under budgeted observational constraints that chooses the next query wisely based on past interactions with BDM. Moreover, we show that we may make even better use of our observations by \emph{exploiting the structure of our problem} through mimicking the BDM.

\textbf{Contributions}\quad We summarize our contributions as follows: 1) we present a simple to implement frequentist approach that enables the use of arbitrary regressors to choose context-action pairs for learning DISS policies;
2) we further introduce and analyze a decision-maker mimicking approach that better utilizes available data; 3) we frame multiple real-world applications within our proposed DISS framework; 4) we empirically show the efficacy of our methods against state-of-the-art methods under various settings and datasets.

\section{Methods}
\subsection{Problem Setting}\label{sec:problem_setting}
Below, we formally introduce our task;
We illustrate this through a running example of clinical decision-making.

\textbf{Decision Maker}\quad
We seek to maximize rewards for the decisions of a BDM $M$ that ingests: 1) arbitrary subsets of available information on an instance $x$; 2) additional options $o$ (both of which will impact $M$'s outputs). To illustrate, a clinician may ingest a given subset of patient features in a visual display style specified by $o$. For notational simplicity, we take $x \in \mathbb{R}^d$. For each instance, we consider $M$'s decision space to be binary for simplicity, $\hat{y}\in \{0,1\}$ (e.g., whether to diagnose a patient with a certain disease). In this case, the BDM $M$ makes decisions based on subsets of features $b \in \{0, 1\}^d$ and additional options $o \in \mathbb{O}$ as $M(x \odot b, o)$\footnote{Note that, depending on the domain of instance features, $x \odot b$ may be ambiguous between missing and zero values. $M$ may equivalently be written as $M(x \odot b + \varnothing(\mathbf{1}-b), o)$ for an empty token $\varnothing$ or as $M(x \odot b, b, o)$. We omit this for notational simplicity.}, where $\odot$ denotes the Hadamard product. We consider the output space of $M$ to be $[0, 1] \times \mathbb{M}$, the tuple of $M$'s probability of selecting $\hat{y}=1$, along with any additional meta-data $m\in \mathbb{M}$ that is relevant for assessing decision quality. For example, a clinician outputs a confidence of their decision (probability of $\hat{y}=1$), along with a recorded time taken to make a decision (the meta-data $m$).

\textbf{DISS Policy and Criterion}\quad
The policy shall receive a reward based on the output of the decision maker and the ground truth optimal decision $y$ for an instance $x$: $r(y, M(x \odot b, o))$. We take the reward function to be known, which reflects typical use cases where the output of the decision-maker, meta-data, and ground truth optimal decisions are deterministically aggregated in some known, application specific manner to measure decision efficacy (e.g., based on accuracy, confidence, speed to decision, etc.).
We wish to learn a policy $\pi: \mathbb{R}^d \mapsto  \{0, 1\}^d \times \mathbb{O}$, that maximizes this reward over a data distribution of instances, $\mathcal{D}$. That is, we want to learn a policy $\pi$ according to:
\begin{equation}
    \label{eq:reward_criteria}
    \argmax_{\pi \in \Pi} \mathbb{E}_{(x,y)\sim \mathcal{D}}[r(y, M(x \odot \pi_b(x), \pi_o(x))],
\end{equation}
where $\pi_b(x),\, \pi_o(x)$ indicate the selected subset of features and options, respectively.
To train the policy, we assume that we have a standard supervised dataset $D = \{(x^{(i)}, y^{(i)})\}_{i=1}^n \overset{iid}{\sim} \mathcal{D}$. Note, however, that $D$ does not contain any information about the decision-making of $M$, which the policy must consider. For many real-world applications, we will be limited on the number of calls that can be made to the BDM $M$ during training. For example, time/incentive limitations on collecting decision-making data from clinicians. Thus, we consider a fixed-budget of $K$ queries to $M$ when training $\pi$.
That is, we consider a paradigm where one sequentially queries the BDM $M$ at $K$ instances, $M(x^{(i_1)} \odot b^{(1)},  o^{(1)}), \ldots, M(x^{(i_K)} \odot b^{(K)}, o^{(K)})$ at select instances $x^{(i_l)} \in D$ and respective information subsets and options $b^{(l)}\in \{0, 1\}^d, o^{(l)}\in\mathbb{O}$. We similarly record rewards $r(y^{(i_1)},M(x^{(i_1)} \odot b^{(1)}, o^{(1)})), \ldots, r(y^{(i_K)}, M(x^{(i_K)} \odot b^{(K)}, o^{(K)}))$. Let
\begin{equation}
    \label{eq:obs_data}
    B^{(K)} \equiv \{ (i_l,  b^{(l)}, o^{(l)}, \eta^{(l)}, \rho^{(l)}) \}_{l=1}^K,
\end{equation}
where $\eta^{(l)} = M(x^{(i_l)} \odot b^{(l)}, o^{(l)}),\, \rho^{(l)} = r(y^{(i_l)}, \eta^{(l)})$, be the dataset of the $K$ acquired decision-making observations.
Since decisions are collected based on historical offline instances $D$, our main concern is not regret at training time.
Rather, we focus on inference time rewards \eqref{eq:reward_criteria}
(see e.g.~\cite{Char2019} for further details on offline multitask reward optimization).

\subsection{Training Observation Acquisition}
\label{sec:trainingacq}
\textbf{Bayesian Regressor-based Policy}\quad Our task connects nicely to Bayesian experimental design and Bayesian Optimization \citep{gree-2020} as we must decide the $K$ decision-making observations to acquire in order to train our policy. Thus, it is natural to consider a Gaussian Process \citep{ling-2016} based reward estimator $f \sim \mathcal{GP}(\mu, \sigma) | B^{(K)}$, which estimates $f(x, b, o) \approx \mathbb{E}_{y \mid x}[r(y, M(x \odot b, o))]$, the expected reward on an instance $x$ with features and options $b, o$ on the BDM $M$ based on observations $B^{(K)}$ \eqref{eq:obs_data}. Let ${\mu}^{(K)}(x, b, o) = \mathbb{E}_{f \sim \mathcal{GP}(\mu, \sigma) | B^{(K)}}[f(x, b, o)]$ be the corresponding posterior mean. Note that one may easily derive a policy based on ${\mu}^{(K)}$ as $\pi(x) \equiv \mathrm{argmax}_{b, o}\, {\mu}^{(K)}(x, b, o)$. Below, we describe strategies to acquire observational data for this policy.

\textbf{Thompson Sampling Acquisition}\quad
Thompson sampling (TS) \citep{thom-1933,thom-1935,russ-2018} provides a principled approach to acquire an additional observation at step $t$ based on the previously acquired data $B^{(t-1)}$ \eqref{eq:obs_data}. \emph{Direct Thompson Sampling} would collect the next observation tuple based on a uniformly sampled instance $i_t \sim \mathrm{Unif}\{1, \ldots, n\}$ and posterior draw $\tilde{f} \sim \mathcal{GP}(\mu, \sigma) | B^{(t-1)}$ using
\begin{equation}
    \label{eq:thompson_add}
    b^{(t)}, o^{(t)} \equiv \mathrm{argmax}_{b, o}\, \tilde{f}(x^{(i_t)}, b, o)
\end{equation}
to update the data as $B{(t)} \equiv \{(i_t,  b^{(t)}, o^{(t)}, \eta^{(t)}, \rho^{(t)})\} \cup B^{(t-1)}$, where $\eta^{(t)}, \rho^{(t)}$ are the corresponding decision-maker output and reward (see \autoref{eq:obs_data}).
\emph{Continuous Multi-Task Thompson Sampling} (CMTS, as in \cite{Char2019}) presents an alternate heuristic, which is able to additionally select the instance $x$ to observe our decision-maker and reward as follows. As before, we draw from the posterior $\tilde{f} \sim \mathcal{GP}(\mu, \sigma) | B^{(t-1)}$. However, CMTS selects
according to improvement:
\begin{equation}
    i_t \equiv \argmax_i  \max_{b,o} \tilde{f}(x^{(i)}, b, o) - \tilde{f}(x^{(i)}, \bar{b}^{(i)}, \bar{o}^{(i)}),
\end{equation}
where $\bar{b}^{(i)}, \bar{o}^{(i)} \equiv \argmax_{b,o}\, \mu^{(t-1)}(x^{(i)}, b, o)$, for posterior mean $\mu^{(t-1)}$,
and proceeds to add an observation as before \eqref{eq:thompson_add}. Both Thompson Sampling approaches rely on a flexible Bayesian regressor, which is commonly taken to be a GP. A slight drawback is that GPs have difficulties scaling to larger datasets and must employ approximations and other optimizations \citep{Murphy2022}. A larger drawback, however, is that GPs often have difficulties modeling higher-dimensional domains \citep{Murphy2022}. For DISS, the regressor must model in the joint space of instance features, subset indicators (which encompass the power set over $d$ elements), and additional display options. Below, we propose an alternative to TS
with a Bayesian model, enabling the use of any regressor.

\textbf{Frequentist Thompson Sampling Acquisition} \quad
As previously noted, one may use bootstrap trials to simulate Thompson Sampling without a Bayesian model \citep{russ-2018,eckl-2014}. Here, we expand on this strategy for DISS by drawing analogues to functional posterior draws and means in a way that enables data acquisition with either TS or CMTS. Bootstrap trials, based on multiple fits to datasets drawn with replacement, measure uncertainty over statistics in a frequentist fashion \citep{dici-1996}. First, we note that under a bootstrap trial with a dataset sampled with replacement, $\tilde{B}^{(t)}$, the fit of a regressor, $f_{\tilde{B}^{(t)}}$, is analogous to a draw from the posterior. Moreover, the posterior mean is then akin to the mean fit over datasets sampled with replacements, $\bar{f} = \mathbb{E}_{\tilde{B}^{(t)}} f_{\tilde{B}^{(t)}}$, which may be approximated using the mean over $C$ datasets. Note that this enables one to use any regressor, including those that scale computationally and are more adept in high-dimensions (e.g., XGBoost \citep{Chen2016a}). We summarize our Frequentist Thompson Sampling (FTS) in \autoref{alg:fts}.
\begin{algorithm}[hbt!]
    \caption{Frequentist Thompson Sampling (FTS)}\label{alg:fts}
    \begin{algorithmic}
        \REQUIRE supervised dataset $D$, budget $K$, warmup budget $t_\mathrm{init}$, ensemble size $C$, batch size $S$.
        \STATE Let $B^{(t_\mathrm{init})} =  \{ (i_l,  b^{(l)}, o^{(l)}, \eta^{(l)}, \rho^{(l)}) \}_{l=1}^{t_\mathrm{init}}$ \eqref{eq:obs_data},
        \STATE \quad where  $i_l,  b^{(l)}, o^{(l)}$ are drawn randomly.
        \FOR{$t = t_{init}+1$ \textbf{to} $K$}
        \STATE Let $F^{(t-1)} = \{ \Tilde{f}_{\Tilde{B}_j}(x, b, o) \}_{j=1}^C$  be regressors fit on
        \vspace{-10pt}
        \STATE \quad $\Tilde{B}_j$ drawn with replacement from $B^{(t-1)}$.
        \STATE Draw $\Tilde{f} \in F^{(t-1)}$, set $\Bar{f} = \frac{1}{C} \sum\limits_{f \in F^{(t-1)}} f$.
        \IF{\texttt{CMTS}}
        \STATE Draw $I = \{i_1, \ldots, i_S\} \overset{iid}{\sim} \mathrm{Unif}\{1, \ldots, N\}$.
        \STATE For $i \in I$ set $\Bar{b}^{(i)}, \Bar{o}^{(i)} = \argmax\limits_{b,o} \Bar{f}(x^{(i)}, b, o)$
        \STATE $i_t \hspace{-1pt} =  \hspace{-1pt} \argmax\limits_{i \in I} \hspace{-2pt} \left[
                \hspace{-1pt}
                \max\limits_{b,o}  \Tilde{f}(x^{(i)}, b, o) \hspace{-1pt}-\hspace{-1pt} \Tilde{f}( x^{(i)}, \Bar{b}^{(i)}, \Bar{o}^{(i)})
                \hspace{-1pt} \right]$
        \ELSE
        \STATE Randomly draw $i_t \sim \mathrm{Unif}\{1, \ldots, N\}$
        \ENDIF
        \STATE Let $(b_t, o_t) = \argmax_{(b,o) \in \{0, 1\}^d \times \mathbb{O}} \Tilde{f}(x^{(i_t)}, b, o)$.
        \STATE Observe $\eta_t = M(x^{(i_t)} \odot b_t, o_t)$, $\rho_t = r(y^{(i_t)}, \eta_t)$
        \STATE Update $B^{(t)} = B^{(t-1)} \cup \{(i_t, b_t, o_t, \eta_t, \rho_t)\}$
        \ENDFOR
    \end{algorithmic}
\end{algorithm}

\subsection{Mimic-Structured Regression}
Thus far, we have formally framed our DISS task as a regression problem and have introduced a frequentist approach that enables us to tune our observation acquisition with arbitrary regressors.
Below, we note a special structure to the DISS regression task and propose to decompose the estimation task into two components that make better use of the data at hand.
Recall that in our DISS framework, the reward function $r$, measuring decision-making efficacy, is known.
We now study the estimated target of our regression task, $f(x, b, o) \approx \mathbb{E}_{y \mid x}[r(y, M(x \odot b, o))]$. Upon inspection, one may note that the target actually depends on two unknown quantities: 1) $p(y \mid x)$, the ground truth conditional likelihood for the optimal decision given $x$; and 2) $M(x \odot b, o)$ the output of the decision-maker at $x$ with a configuration $b, o$.
This observation, which has not been previously used in contextual bandit decision support literature (to the best of our knowledge), enables us to better leverage our data. Namely, $p(y \mid x)$ \emph{is independent of the blackbox decision-maker}, and can directly be estimated in a supervised fashion given our dataset of instances $D = \{(x^{(i)}, y^{(i)})\}_{i=1}^n$. That is, we propose to directly estimate $\hat{p}(y \mid x)$ (a classifier in the discrete $y$ case), which avoids any budget expenditure
on observations of $M$.
$M(x \odot b, o)$, on the other hand, depends on the decision-maker and must be estimated with budgeted observations. We propose to learn to \emph{mimic} the BDM based on the observations in $B^{(T)}$ \eqref{eq:obs_data} in a supervised fashion:
\begin{equation}
    \hat{M} \equiv \argmin_{M^\prime} \sum_{t=1}^{T}  \ell( M^\prime(x^{(i_t)} \odot b^{(t)}, o^{(t)}), \eta^{(t)} ),
\end{equation}
Where $\ell$ is a supervised loss (e.g., cross-entropy), and $\eta^{(l)} = M(x^{(i_l)} \odot b^{(l)}, o^{(l)})$ are the observed decisions (and meta-data) from the BDM. With both $\hat{p}(y \mid x)$ and $\hat{M}$ we can directly estimate rewards using the mimic-structured regressor:
\begin{equation}
    \hat{f}_\mathrm{MS}(x, b, o) \equiv \mathbb{E}_{y \sim \hat{p}(y \mid x)}[r(y, \hat{M}(x \odot b, o))].
    \label{eq:MSreg}
\end{equation}
Please see the \autoref{sec:appx_proof} where we expound on a risk upper-bound and on the efficiency of the mimic-structured estimator, which essentially estimates two classifiers (one of which, $\hat{p}(y \mid x)$ is fitted on un-budgeted data) versus the regression approach which can be viewed as fitting either $2^d$ regressors in $d$ dimensions, or as fitting in a regressor in a larger $2\!\cdot\! d$ space.
Moreover, we may still use FTS  (\autoref{alg:fts}) by estimating $\hat{p}(y \mid x)$ and $\hat{M}$ with respective bootstrap trials, which yields a natural way of acquiring data to resolve ambiguities between our knowledge of what the BDM will do and the impact on rewards.

\section{Applications}
\label{sec:applications}
Next, we show how to formally frame several applications within our DISS framework.

\textbf{DISS for Prediction Support} \quad We focus on training a policy to identify a per-instance configuration (subset of features, $b$, and options, $o$) to forward to the decision maker as it makes its prediction (e.g., classification).
The inputs that the decision-maker receives are: 1) a subset of the information on an instance (as indicated by the masked feature vector $x \odot b$); 2) additional options, $o$ (represented in vector form, if applicable).
In the binary prediction case, the decision-maker outputs a class probability (we omit meta data outputs for simplicity below). It is natural to consider a reward based on the log-likelihood of the prediction, $\eta = M(x \odot b, o)$  for the true class, $y$:
\begin{align}
    r(y, M(x \odot b, o)) =
    y \log{\eta} + (1-y) \log({1- \eta}).
    \label{eq:base_reward}
\end{align}

\textbf{Expert Selection}\quad Next, we show how DISS enables selection among multiple experts through an option that policy determines. In this case, we write the BDM as:
$M\left(x \odot b, (j, o) \right) = M_j(x \odot b, o)$,
where the option $j$ indexes which of the available decision-makers ($M_1, \ldots M_J$) to apply to an instance with subset $b$ and further options $o$. Other details may otherwise remain the same as in the single decision-maker case.



\textbf{LLM Decision-Making} \quad
We now consider a BDM \textit{M} that is a fixed, pretrained large language model (e.g., GPT-4 \citep{openai2024gpt4technicalreport}, LlaMA \citep{touvron2023llamaopenefficientfoundation}).
The LLM makes predictions by outputting parsable text containing labels and confidence about given instances based on input prompts (style, content, and subset of features) that are adjusted in a per-instance fashion according to the policy. As aforementioned, many LLMs (and their efficacy) are sensitive to prompting choices \citep{liu-etal-2024-lost, Zhao2021CalibrateBU, anagnostidis2024susceptiblellmsinfluenceprompts}.
To maximize the LLM's per-instance prediction efficacy, we serve the LLM a prompt (or input text) $\texttt{prompt}(x \odot \pi_b(x), \pi_o(x))$ based on the DISS policy, where $\pi_b(x)$ and $\pi_o(x)$ represent the selected feature subset and option for any given instance $x$. For example, in healthcare applications, for a patient $x$, $\pi_b(x)$ controls the set of patient attributes to be presented, and $\pi_o(x)$ controls the style (preamble, wording, etc.) of the prompt containing the chosen patient attributes forwarded to the LLM to obtain an accurate diagnostic prediction in the text.

\textbf{Interpretability}\quad
The interpretability of ML predictions is hindered with larger instance dimensionalities \citep{kari-2023}. We can utilize DISS to encourage interpretability as follows. First, a cardinality penalty may be added to any reward, $r$, to promote efficacious decisions with fewer features: $r^\prime(y, M(x \odot b, o)) = r(y, M(x \odot b, o)) - \lambda \|b\|_1$. Moreover, we can further enforce interpretability by explicitly restricting the action space of the DISS policy to only consider subsets of a small dimensionality (e.g., 3: $\{b \mid \|b\|\leq 3\}$). Lastly, applying DISS to a decision-maker that is itself interpretable further yields interpretable predictions. For example, a localized linear model fitted to the provided query features $x \odot b$ using neighbors in the dataset $D = \{(x^{(i)}, y^{(i)})\}_{i=1}^n$ would directly provide an interpretable decision-making process (a linear combination of a few features fit on neighbors) and gives a complementary method to local interpretable model-agnostic explanations (LIME) \citep{ribe-2016}, or gradient approaches (e.g., \cite{wang-2024}).

\section{Related Work}
\textbf{Contextual Bandits} \quad
DISS may be framed as a contextual multi-armed bandit problem \citep{lu-2010}, where the context is an instance $x \in \mathbb{R}^d$, the action space are all the possible subsets of $d$ features (and a finite set of options), and rewards stem from the BDM's prediction quality when given the subset of features (and options) from the context. (Please see \cite{Lattimore2020} for a general survey on contextual bandits.) Note that the cardinality of the action space $\left\lvert \mathcal{A} \right\rvert \geq 2^d$ grows exponentially in the number of features, making for a difficult contextual bandit setting, which itself is an active area of research (e.g., \citep{Gai2012,Chen2013,Chen2016,Wang2018}).
\cite{Char2019} proposes `Offline Contextual Bayesian Optimization' where one can select contexts during training to observe in order to maximize multi-task inference-time rewards, which fits our DISS framework as discussed in \autoref{sec:trainingacq}. However, their approach, CMTS, operates over with GP regressors, which often have difficulties modeling higher-dimensional domains \citep{Murphy2022} and would have to operate over a $2 \hspace{-1pt}\cdot\hspace{-1pt} d$ joint space (of features and masks). Thus, we alleviate this issue with our frequentist Thompson sampling approach (\autoref{alg:fts}).

\textbf{Adaptable Displays}\quad
The problem of providing decision support via adaptive interfaces has been studied in the context of human-computer interaction (HCI).
\cite{wang_adaptive_2024} surveyed adaptive user interfaces for long-term health patients with predictive algorithms (e.g. reinforcement learning, Monte Carlo tree search).
Typically previous approaches consider a limited number of options to adjust interfaces for human decision-making.
For example, \cite{buçinca_offline_2024} explored offline reinforcement learning to learn policies that adjust four options (vary the type of AI prediction assistance to provide) to aid human learning of tasks.
\cite{lomas_bandit_2016} employed multi-armed bandits to optimize the experience of an online educational game by selecting among six interface variants.
Modiste \citep{Bhatt2023} employed variants of UCB \citep{guan-2018} to learn a policy used to assist human decision-maker's accuracy through providing personalized additional relevant information (over three options) such as LLM output.
DISS, in contrast, selects over a much larger number of adjustments than previous approaches, and also encompasses many more applications beyond human-assistance.

\textbf{Prompt Engineering LLMs}\quad
LLMs can adapt to a myriad of tasks with adequate prompting
\citep{Radford2019LanguageMA}.
Recent work has studied various prompting heuristics, including in-context learning \cite{few-shot}, chain-of-thought prompting \cite{wei2022finetuned} and prompt-tuning \cite{lester-etal-2021-power}.
However, improper prompting may cause the LLM to output poor responses in deployment \cite{Burnell2023-sn}.
Moreover, recent works also show that the LLM outputs are sensitive to the prompting style, position of information in the prompt, and the length of the prompt \citep{liu-etal-2024-lost,Zhao2021CalibrateBU,anagnostidis2024susceptiblellmsinfluenceprompts}.
\cite{zollo2023prompt} introduces an approach to select prompts from a static set based on statistical upper bounds to avoid unacceptable outcomes generated by the LLM. However, this line of work is limited to the expertise of the prompt designer to create a relatively small fixed set of candidate prompts.
Instead, our approach presents an alternate formulation that encompasses such prompting style choices along with a per-instance tailored curation of information to forward to an LLM.

\section{Experiments}
We conduct an extensive set of experiments to test our methodology utilizing various BDMs and datasets. Methods were tested in applications to: 1) aid biased decision-makers; 2) select among multiple decision-makers; 3) build interpretable predictions; 4) aid LLM-based decision-makers. We used various publicly available datasets from UCI ML repo.~including: Skin Segmentation (`skin') \citep{bhat-2009}, Statlog Shuttle (`space') \citep{hene-1992}, SUSY \citep{whit-2014}, Diabetes (`diabetes') \citep{diabetes} and California Housing \citep{housing} (`housing') (see \autoref{sec:appx_dataset} for details). Across applications and datasets, our experiments span \emph{17 environments} to provide an extensive evaluation on methods.

Unless otherwise noted, we use negative cross-entropy loss (label likelihood) between the ground truth label and the decision-maker's prediction
as the reward signal to maximize.
We note that the large actions space (which encompasses the power set over dimensions), makes it challenging to scale most reward maximization approaches.
We compare primarily to concurrent work, Modiste \citep{Bhatt2023}, which considers KNN-UCB \citep{guan-2018} contextual bandit scheme to aid human decision makers. Note, however, the original Modiste paper only considers a few actions (typically three) to customize data fed to decisions. Moreover, Modiste assumed the reward to be bounded and discrete; yet, our reward is continuous and unbounded. As workaround, we make two attempts: the first approach, \texttt{Modiste-KNN}, follows the original Modiste formulation, but we initialize previously unseen reward to be the minimum reward stored in the replay buffer; the second approach, \texttt{Modiste-UKNN}, concatenates feature mask to context and performs k-nearest neighbor regression in this space (see \autoref{sec:appx_modiste_impl}). For further context, we also used continuous multitask Thompson sampling (CMTS) \citep{Char2019} with our proposed frequentist continuous multitask Thompson sampling (\texttt{FCMTS}) routine that utilizes XGB Regressors \citep{Chen2016a} for better scaling; \texttt{Random} utilizes the XGB Regressor on randomly collected decision-making data \eqref{eq:obs_data}. We compare to using Thompson Sampling with our \texttt{Mimic} structured regressor approach \eqref{eq:MSreg}.\todo{\texttt{Mimic} is structured with our proposed FTS not plain thompson sampling. Not sure if this is correct.} All methods are seeded with an initial set ($t_\mathrm{init}$, \autoref{alg:fts}) of $500$ observations. We plot held-out test time rewards vs number of acquired observations to report policies' inference time performance under various budgets. Code will be open-sourced upon publication.
Run-times across acquisition approaches were comparable; wall-clock on the `susy' environment ran on an Nvidia A100 partition was 16.13, 14.55, and 19.44 hours for \texttt{Mimic}, \texttt{FCMTS}, \texttt{Modiste KNN} respectively (see \autoref{sec:appx_experimental} for further experimental details).

\subsection{Biased Synthetic Expert Setting}
\label{biased_synth_exp}

Motivated by the target application of providing decision support to human experts, we have devised a set of experiments featuring a synthetic decision maker exhibiting various forms of simulated cognitive bias. We utilize a Nadaraya-Watson label smoother (see \autoref{sec:appx_experimental_synthetic_expert})
that can make predictions given arbitrary subsets as a base decision-maker that is biased under various conditions, as follows.

\textbf{Cognitive Overload}\quad Human are often hampered by the limited ability of their working memory to hold multiple simultaneous pieces of information. When these limits are reached, one is likely to experience ``cognitive overload,'' making it difficult to predict accurately and with confidence \citep{iskander_burnout_2018}.
We simulate the effects of cognitive overload by applying a temperature-controlled softmax activation function to the label probabilities
obtained from our synthetic expert:
$\text{softmax}(\mathbf{\hat{y}}, {b})_i = \frac{\exp{(\frac{\hat{y}_i}{T({b})}})}{\sum_{j=1}^{\text{nclass}}\exp{(\frac{\hat{y}_j}{T({b})}})}$,
where $T({b})$ is a temperature function whose value increases as more features are observed in ${b}$ (we set $T({b})$ to be proportional to the 2-norm of ${b}$).

\begin{figure}[h]
    \centering
    \includegraphics[width=\linewidth]{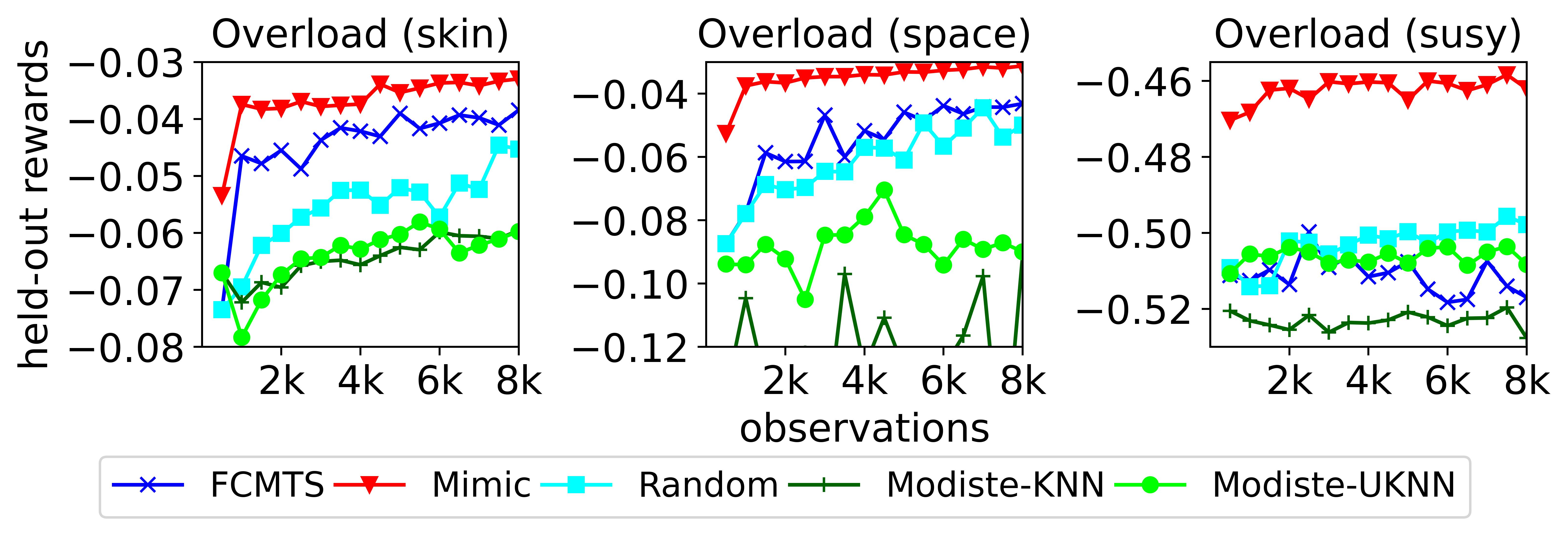}
    \caption{Average rewards vs.~observation data budget on cognitive-overloaded expert environment.
    }
    \label{fig:overload}
\end{figure}

In order to provide effective decision support to this type of BDM, we expect our policy to learn a small but efficient feature subset that allows accurate prediction while incurring minimal uncertainty penalties. In \autoref{fig:overload}, we see that in this setting, our \texttt{Mimic} policies handily outperform other baselines.
In fact, the maximum \texttt{Modiste} reward threshold was already exceeding utilizing our \texttt{Mimic}-structured regressor after the warm-up period of 500 observations, further illustrating the power of exploiting the DISS structure.

\paragraph{Risk Aversion} Experts may be biased due to out-sized risk aversion;  e.g., in medical settings \citep{sorum_risk_2003}
or loss avoidance in international trade \citep{tovar_aversion_2009}.
To simulate risk aversion (to false negatives), we combine the decision-maker's prediction with a 100\% positive prediction, 
weighted by a bias strength parameter $\alpha$:
$ \hat{y}_{\text{averse}} = (1-\alpha) \hat{y} + \alpha (1)
$.

\begin{figure}[h]
    \centering
    \includegraphics[width=\linewidth]{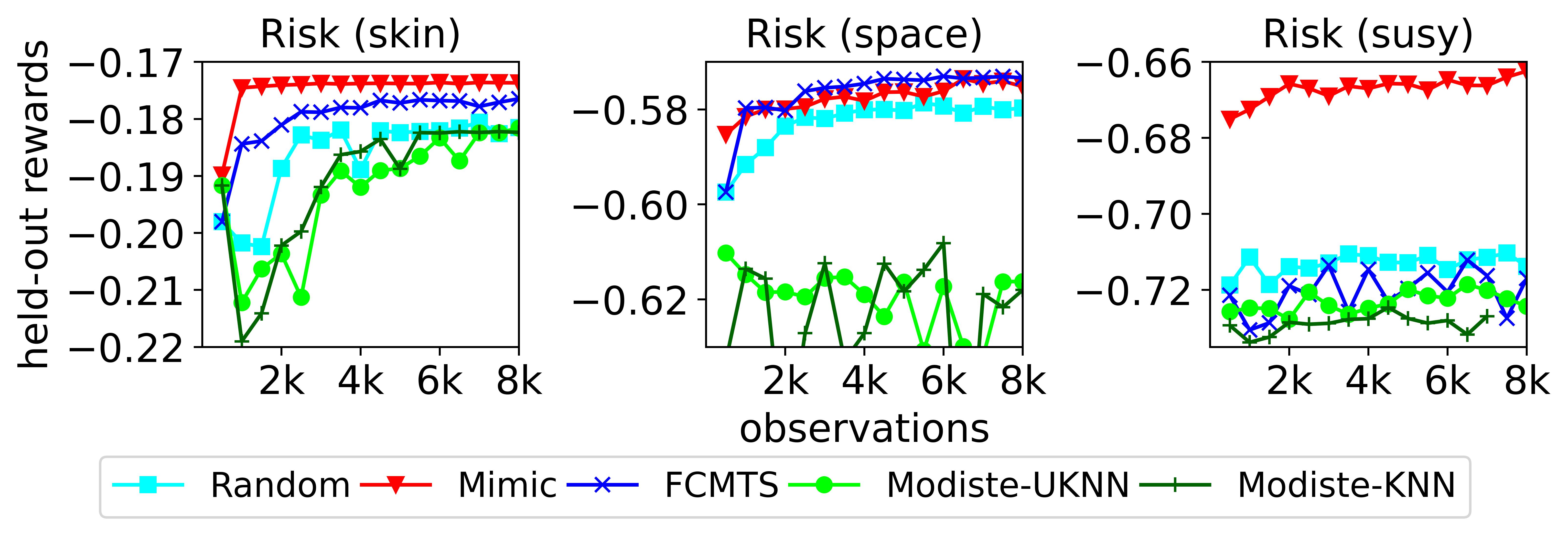}
    \caption{
        Average rewards vs.~observation data budget on risk-averse expert environment (overly concerned with minimizing false negative predictions).
    }
    \label{fig:risk_averse}
\end{figure}

To successfully support this BDM's decision making, we expect our policy to select features that increase the BDM's confidence when predicting negative examples.
We see similar trends in \autoref{fig:risk_averse} in that \texttt{Mimic} policies quickly converge to a high reward value.

\paragraph{Simplicity Bias}\quad
Rather than viewing data profile holistically, humans may allocate undue attention towards prejudicial factors or simple explanations stemming from a few features, such as in the medical field where racial and gender biases are known to lead to discrepancies in treatment recommendations \citep{chen_racial_2001, schulman_effect_1999}.
We model simplicity bias in prediction by computing a convex combination between the BDM's original prediction and a simplified prediction. In this case, the biased prediction comes from a univariate classification model $g(x_j)$ trained only on the $j^{th}$ feature column of the data (the ``poison feature.'').
If the poison feature is selected in ${b}$, the expert's original prediction will be modified as
$\hat{y}_{\text{poison}} = (1-\alpha) \cdot \hat{y}+ (\alpha) \cdot g(x_j)$
Otherwise, the prediction will be unchanged.

\begin{figure}[h]
    \centering
    \includegraphics[width=\linewidth]{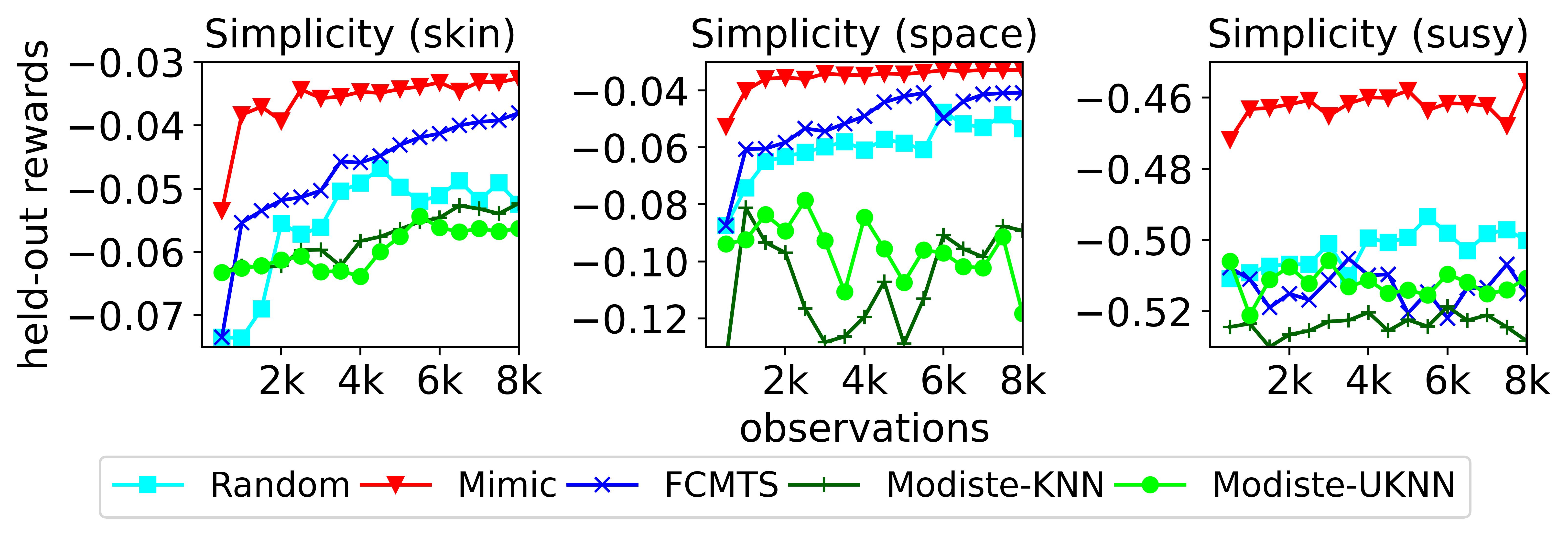}
    \caption{
        Average rewards vs.~observation data budget on oversimplification expert environment.
    }
    \label{fig:poison}
\end{figure}
To avoid the effects of the simplification bias, we expect our method to learn a policy that avoids including the poison feature in the subsets presented to the expert.  \autoref{fig:poison} showcases
that our \texttt{Mimic} based policies are also successful in this environment.

\textbf{Results}\quad We observe that on the higher-dimensional datasets (`space' and `susy'), \texttt{Modiste} struggles to consistently increase its rewards as more examples are observed.
While the more sophisticated regressors in \texttt{FCMTS} aid performance, we consistently see the best rewards from our proposed \texttt{Mimic} structured approach.

\subsection{Multi-Expert Setting}

Next, we consider a scenario where multiple decision-makers are available to the policy, but only one can be queried due to constraints. Thus, in addition to selecting a subset of features to display, the policy must also select what decision-maker to utilize.
In real-world settings, decision makers may have varying expertise on different instance sub-spaces. We simulate this property by separating our data into clusters and providing each expert with a single cluster as the support for their knowledge base. For our experimental trials, we performed K-Means clustering on each dataset with $k=4$, resulting in four distinct experts for our policy to choose from. Unlike the biased synthetic expert setting in \autoref{biased_synth_exp}, synthetic experts in these experiments the were not modified with any additional source of limitation or bias; only the source of the models' training data differed between experts (see \autoref{sec:appx_experimental_multi_expert}).
\begin{figure}[h]
    \centering
    \includegraphics[width=\linewidth]{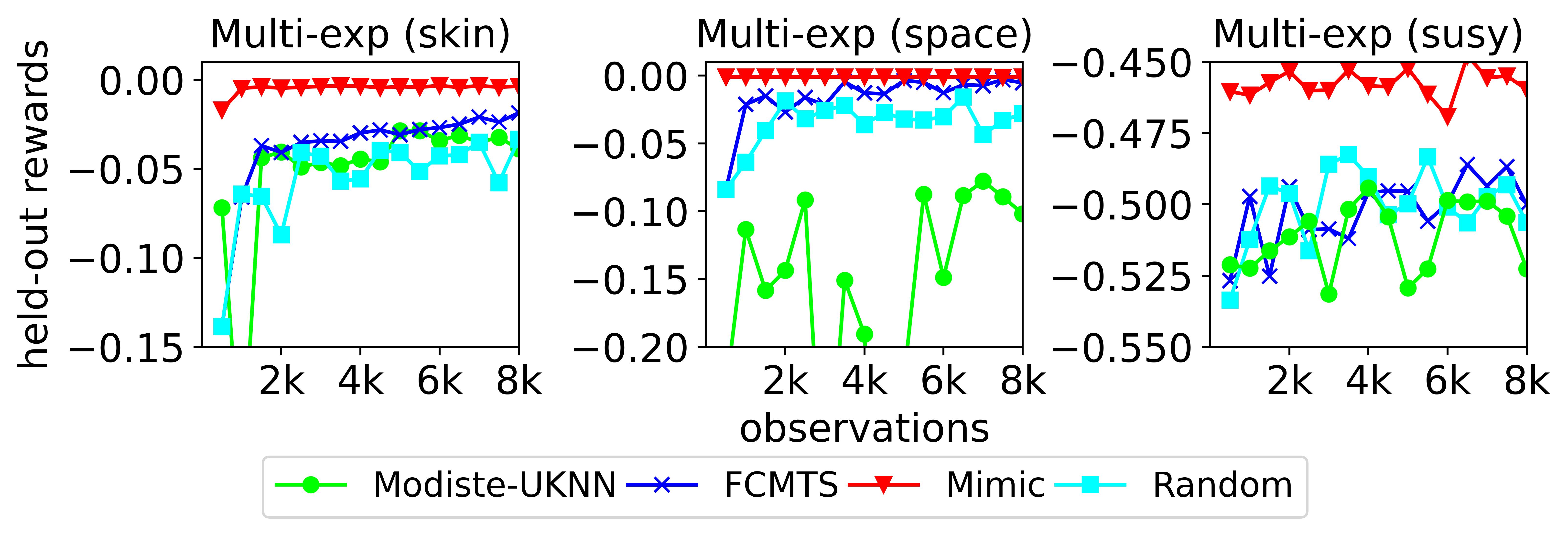}
    \caption{
        Average rewards vs.~observation data budget on the multi-expert environment. (\texttt{Modiste-KNN} failed to produce rewards within reported ranges.)
    }
    \label{fig:multi}
\end{figure}

The results in \autoref{fig:multi} indicate that \texttt{Modiste} policies with a unified-KNN reward estimator can compete with \texttt{FCMTS} policies with an XGB reward estimator (except in the case of the space shuttle dataset); however, \texttt{Mimic} policies utilizing
consistently outperform both. In fact, the reward earned by latter reaches the theoretical maximum 0.0 for the two lower dimensional datasets, suggesting that having the option to select among different decision makers allows for better overall predictions compared to relying on a single expert.

\subsection{Interpretability and Visualization }\label{sec:exp_interpretability}
\begin{figure}[h]
    \centering
    \includegraphics[width=\linewidth]{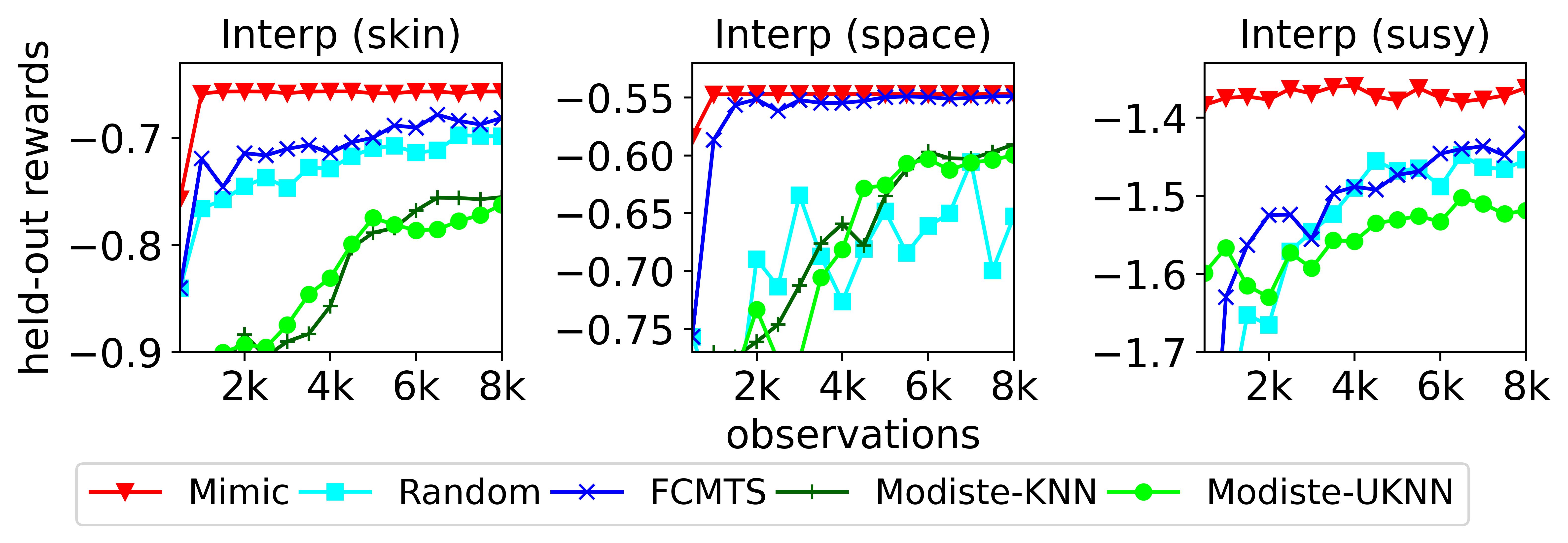}
    \caption{
        Average rewards vs.~observation data budget on interpretable prediction environment with $\lambda=0.5$.
    }
    \label{fig:beta_sweep_curves}
\end{figure}
\begin{wrapfigure}{r}{0.5\linewidth}
    \centering
    \includegraphics[width=\linewidth]{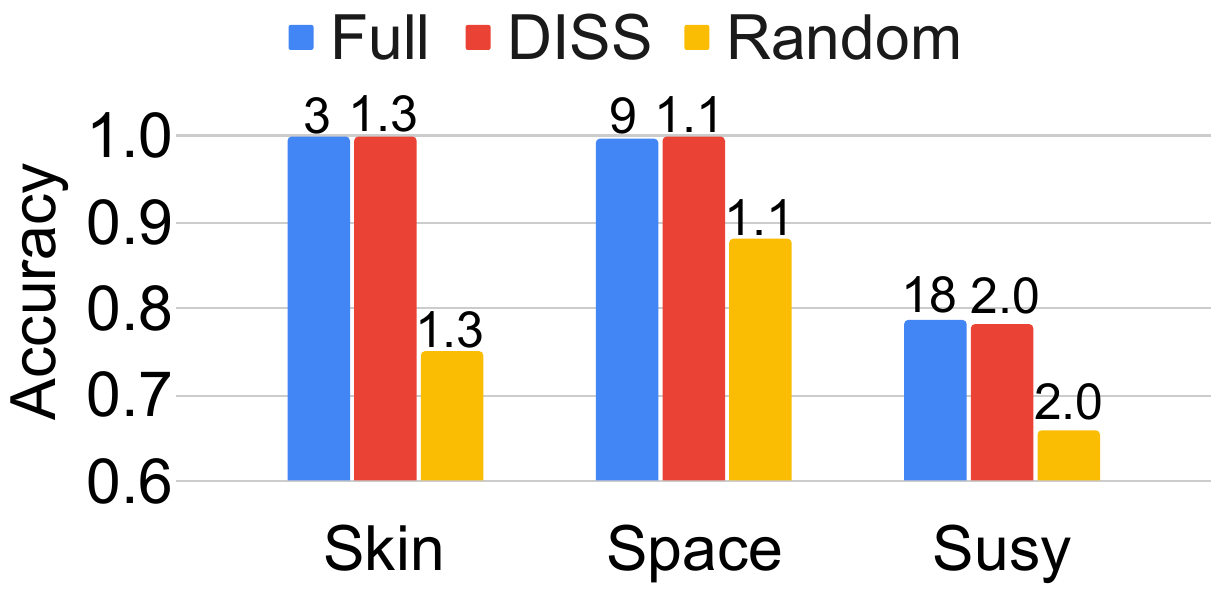}

    \vspace{6pt}
    \includegraphics[width=\linewidth]{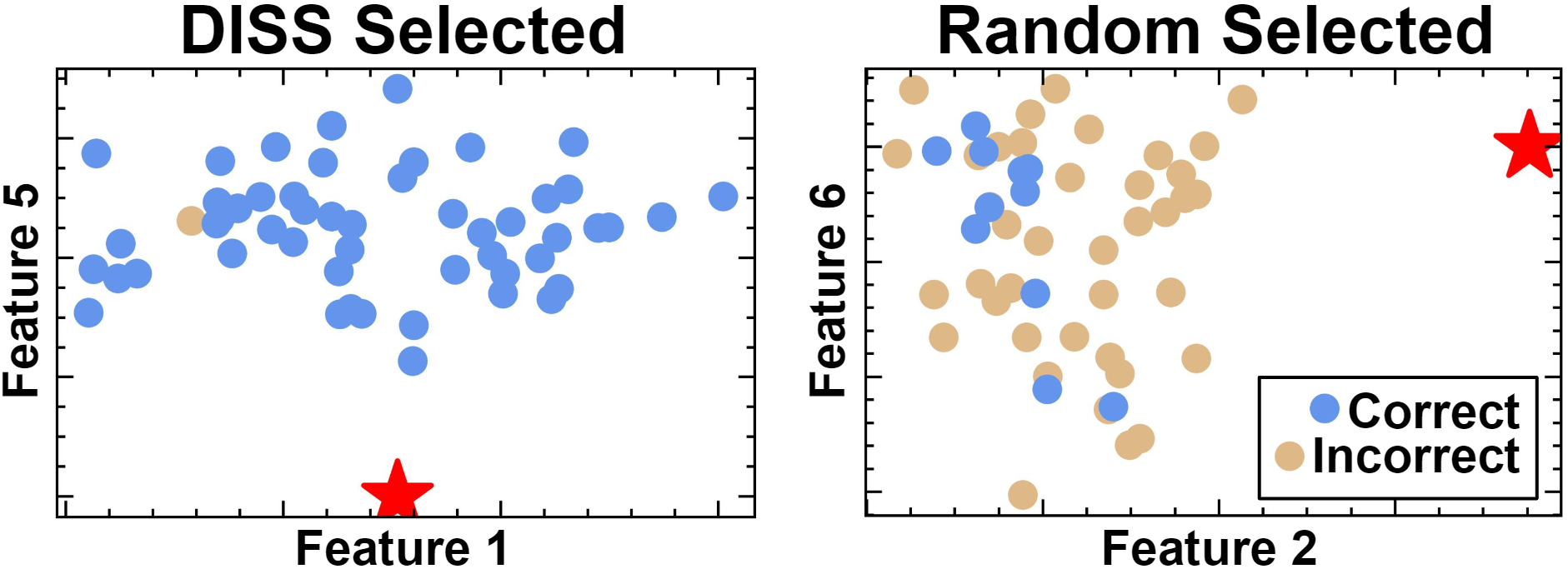}
    \caption{\emph{Top}: Accuracy of \texttt{Mimic} selected features vs.~random features and full feature classification  (average selected reported).
        \emph{Bottom}: Scatter plot of \texttt{Mimic} selected features.
    }
    \label{fig:space_interp}
\end{wrapfigure}
Next, we provide a proof-of-concept application of DISS to yield more interpretable predictions. In \autoref{sec:applications}, we discussed a penalty term that subtracts the number of features selected weighted by a hyperparameter $\lambda$ to encourage the policy to use a more compact subset of features. Here our base decision-maker is a linear (logistic regression) classifier, trained on neighbors from the forwarded $x \odot b$ features on the training set. That is, the BDM returns predictions from a locally trained linear model on the query $x \odot b$. The cost incurred from each call (stemming from a neighbor search and model training) and the simplicity of the final prediction (a linear combination of selected features) makes this environment compelling as a use case (see \autoref{sec:appx_experimental_vis_and_interp}).

In \autoref{fig:beta_sweep_curves}, we again see that our \texttt{Mimic} approach typically reaches its top performance faster than \texttt{FCMTS},
which verifies our intuition regarding taking advantages of our prior knowledge to the reward signal and therefore has better sample efficiency than its counterparts.
Furthermore, we see in \autoref{fig:space_interp} (top), that our \texttt{Mimic} policies are surprisingly adept at uncovering a dynamic subset of features to predict on, and can achieve comparable accuracies to a full model trained with all feature using two or fewer features on average with a personalized linear model.
Moreover, we visualize the typical neighbors in the selected features in the `space' dataset in \autoref{fig:space_interp} (bottom), and see much more structure (and correct class predictions) in selected features than with random feature sets.

\subsection{Large Language Model Setting}\label{sec:llm_setting}

\begin{wrapfigure}{r}{0.5\linewidth}
    \centering
    \includegraphics[width=\linewidth]{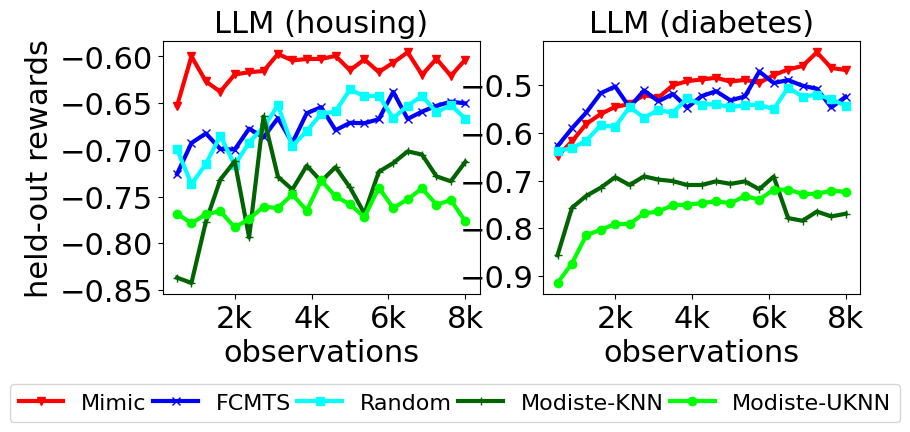}
    \caption{Average rewards vs.~budget for LLM BDMs.}
    \label{fig:llm-result}
\end{wrapfigure}
LLMs are effective in distilling common-sense knowledge about various domains, making them appealing as general decision-makers \cite{yao2023treethoughtsdeliberateproblem}.
Unfortunately, LLMs are sensitive to how input text is structured \citep{liu-etal-2024-lost,Zhao2021CalibrateBU,anagnostidis2024susceptiblellmsinfluenceprompts}. Thus, here we study DISS's ability to provide decision support for black-box LLMs to make decisions on unseen data. In particular, we consider the Diabetes dataset \citep{diabetes} and California Housing Dataset \citep{housing}, which contain features that typically LLMs can reason over. We used \texttt{Meta-Llama-3.1-8B-Instruct-Turbo} as the LLM since it achieves strong performance on a number of benchmarks \citep{dubey2024llama3herdmodels}. As discussed in \autoref{sec:applications}, for each instance, we utilize policies to determine what feature to include in the input text that is presented to the LLM for predictions about the target label (see \autoref{sec:appx_experimental_llm}).
Interestingly, we observe that we can achieve better output predictions and confidences by strategically withholding certain information from the LLM in a per-instance fashion (see \autoref{sec:appx_llm_examples}). Again, we see that our proposed \texttt{Mimic} approach has the best performance.

\section{Conclusion}
In this work, we present a novel AI assistance framework -- Dynamic Information Sub-Selection (DISS) -- to enhance the performance of black-box decision-makers by dynamically tuning the presented information. We develop a frequentist approach for DISS policies to acquire data that enables the use of arbitrary regressors; moreover, we propose a novel mimicking approach that takes advantage of the structure of the DISS regression task to better utilize the data at hand. We show how various real-world applications (including biased decision support, expert selection, interpretable modeling, and LLM support) can be framed under our DISS framework. Extensive experimentation shows our proposed \texttt{Mimic} approach achieves state-of-the-art across these applications.
We believe that the success of our \texttt{Mimic} based policies in these experiments can be used as a stepping stone for further development across these applications in future work.
\printbibliography[segment=0,heading=subbibliography]
\onecolumn
\aistatstitle{Dynamic Information Sub-Selection for Decision Support:\\Supplementary Material}
\appendix
\newrefsegment
\setcounter{equation}{0}
\setcounter{figure}{0}
\setcounter{table}{0}
\setcounter{algorithm}{0}
\section{Mimic-Structured Regressor Analysis}\label{sec:appx_proof}
Below we provide further context for the behavior of the mimic-structured regressor under a simplified binary setting where: 1) the BDM does not take in options and may be written as $M(x \odot b)$; 2) the BDM only outputs a probability of selecting $y=1$, $M: \mathbb{R}^d \mapsto [0,1]$; 3) the reward function is bounded by a constant, $|r(y, M(x \odot b))|\leq R$; 4) the reward function is Lipschitz smooth $|r(y, \eta)-r(y, \eta^\prime)|\leq L |\eta - \eta^\prime|$, for $y\in\{0,1\}$. Then, for all $x, b$, we have (using the triangle inequality):
\begin{align*}
     & \left| \mathbb{E}_{p(y \mid x)}\left[r(y, M(x \odot b))\right] - \mathbb{E}_{\hat{p}(y \mid x)}\left[r(y, \hat{M}(x \odot b))\right] \right|                 \\
     & \leq  \left|p(y=1 \mid x) r(1, M(x \odot b))- \hat{p}(y=1 \mid x) r(1, \hat{M}(x \odot b))\right|                                                            \\
     & \quad +\left|p(y=0 \mid x) r(0, M(x \odot b))-\hat{p}(y=0 \mid x) r(0, \hat{M}(x \odot b)) \right|                                                           \\
     & \leq  \left|p(y=1 \mid x) r(1, M(x \odot b))
    - \left(\hat{p}(y=1 \mid x)+p(y=1 \mid x)-p(y=1 \mid x)\right) r(1, \hat{M}(x \odot b)
    )\right|
    \\ &\quad + \left|p(y=0 \mid x) r(0, M(x \odot b))
    -\left(\hat{p}(y=0 \mid x)+p(y=0 \mid x)-p(y=0 \mid x)\right) r(0, \hat{M}(x \odot b)
    ) \right|                                                                                                                                                       \\
     & \leq \Big|p(y=1 \mid x)\big(r(1, M(x \odot b))-r(1, \hat{M}(x \odot b))\big)\Big| +\Big|\big(p(y=1 \mid x)-\hat{p}(y=1 \mid x)\big) r(1, \hat{M}(x, b))\Big| \\
     & \quad +\Big| p(y=0 \mid x) \big(r(0, M( x \odot b))-r(0, \hat{M}(x \odot b)\big) \Big| +|(p(y=0 \mid x)-\hat{p}(y=0 \mid x)) r(0, \hat{M}(x \cdot b))|       \\
     & \leq  2L\big|M(x \odot b)-\hat{M}(x \odot b)\big|+\Big(|p(y=1 | x)-\hat{p}(y=1 \mid x)|+|p(y=0 \mid x)-\hat{p}(y=0 \mid x)|\Big)R.
\end{align*}
Then, further supposing: 1) $X$ is bounded a.s.; 2) $p$ is Lipschitz smooth, $|p(y \mid x) - p(y \mid z)| \leq C \|x-z\|$ for all $y,x,z$, and following \cite{gyorfi2007nonparametric}, we may bound
\begin{align*}
    \mathbb{E}\left[|p(y=1 | x)-\hat{p}(y=1 \mid x)|+|p(y=0 \mid x)-\hat{p}(y=0 \mid x)|\right] \leq c\, n^{-\frac{1}{d+2}},
\end{align*}
for a constant $c$ using a non-parametric partitioning estimate for $\hat{p}$ on an \emph{iid} dataset $D = \{(x^{(i)}, y^{(i)})\}_{i=1}^n$ (see Thm.~2 \citep{gyorfi2007nonparametric}).
Similarly (with accompanying Lipschitz assumptions on $M$),
\begin{align*}
    \mathbb{E}\left[|M(x \odot b)-\hat{M}(x \odot b)|\right] \leq c^\prime\, t^{-\frac{1}{d+2}},
\end{align*}
by using a respective non-parametric partitioning estimates for $\hat{M}$  on an \emph{iid} dataset $B = \{(x^{(j)} \odot b^{(j)}, \eta^{(j)})\}_{j=1}^t$, $\eta^{(j)}\sim M(x^{(j)} \odot b^{(j)} )$.
Taking $t \in \mathcal{O}(n)$,
we may then simplify:
\begin{align*}
    \mathbb{E}\left[\left| \mathbb{E}_{p(y \mid x)}\Big[r(y, M(x \odot b))\Big] - \mathbb{E}_{\hat{p}(y \mid x)}\left[r(y, \hat{M}(x \odot b))\right] \right|\right] \leq c^{\prime\prime}\, t^{-\frac{1}{d+2}}.
    \qed
\end{align*}
We see that the error (on randomly collected data) then is dominated by the number of budgeted observations on the BDM, $t$. 
Furthermore, one may compare to a standard regression approach to estimate rewards, which must operate in the joint $2\!\cdot\! d$ space, $\hat{f}(x,b)\approx \mathbb{E}_{p(y \mid x)}\left[r(y, M(x \odot b))\right]$, where corresponding mean absolute error bounds for partition estimators yield a slower rate of $t^{-\frac{1}{2d+2}}$ (see Chapter 4.4 \citep{gyorfi2006distribution}). Alternatively, viewing the regression task as approximating $2^d$ separate regressors in $d$ dimensional space $\hat{f}_b(x)\approx \mathbb{E}_{p(y \mid x)}\left[r(y, M(x \odot b))\right]$ would place an exponential (in $d$, data dimensionality) factor on the required number of budgeted acquisitions.

\section{Dataset}\label{sec:appx_dataset}
Statistics of the dataset used in our experiments are shown in \autoref{tab:dataset}. The skin segmentation, statlog (shuttle), and SUSY datasets are used in biased synthetic expert, multi-expert, and interpretable settings, whereas pima diabetes and california housing datasets are used in the large language model setting.
\begin{table}[h]
    \centering
    \caption{Dataset Information} \label{tab:dataset}
    \begin{tabular}{lcc}
        \toprule
        \textbf{Dataset}   & \textbf{Instances} & \textbf{Features} \\
        \midrule
        Skin Segmentation  & 245057             & 3                 \\
        Statlog (Shuttle)  & 58000              & 7                 \\
        SUSY               & 5000000            & 18                \\
        Pima Diabetes      & 10000              & 8                 \\
        California Housing & 20640              & 8                 \\
        \bottomrule
    \end{tabular}
\end{table}

\section{\texttt{Modiste} Implementation}\label{sec:appx_modiste_impl}
There are two obstacles when apply \texttt{Modiste} KNN variant (see Alg. 3 in \citep{Bhatt2023}) to our DISS framework, to which we provide two workarounds termed \texttt{Modiste-KNN} and \texttt{Modiste-UKNN}.

\subsection{\texttt{Modiste-KNN}}\label{sec:appx_modiste_knn}
\cite{Bhatt2023} assumed the reward to be bounded and discrete, specifically $r: \mathcal{X} \times \mathcal{A} \rightarrow \left\{ 0, 1\right\}$. Meanwhile, to estimate reward for given any context-action pair, the original $\texttt{Modiste}$ KNN variant queries the nearest neighbors in the context space from replay buffer that has the same action with the given input (see Alg. 3 step 1 in \cite{Bhatt2023}); yet, as the size of action space increases, it is more than likely not every action gets observed during training time; for these unobserved actions, the original \texttt{Modiste} KNN variant used the fact that reward is either zero or one and set the estimated reward to the default value of 0.5 if the queried action is not observed during training. However, we do not have a simple analogous counterpart to their default value of 0.5 since our reward is unbounded. Hence, for actions that were not observed during training, we initialize it to the minimum reward observed in the current replay buffer. We note that other potential options for statistics to consider include mean and $q^\text{th}$-quantile.

\subsection{\texttt{Modiste-UKNN}}
As size of action space increases, the possibility of the score estimator not able to acquire sufficient interactions with the environment increases. The workaround we proposed in \autoref{sec:appx_modiste_knn} only addresses the issue of which default value to use for unobserved action but did not tackle the fact that a default value is needed for unobserved actions. Here, we note that it is highly likely that DISS actions are related, i.e. the action that uses feature 6 and 7 could perform well if we already have prior experience taking actions that use feature 7. Based on this intuition, we perform k-nearest neighbors query directly in context-action space with the following customized distance metric:
\begin{equation}\label{eq:appx_modiste_uknn_dist}
    \text{distance}(\langle x, a \rangle, \langle x^\prime, a^\prime \rangle) = \left\lVert x, x^\prime \right\rVert + \nu \left\lVert a, a^\prime \right\rVert
\end{equation}
where $\nu$ is a hyperparameter. In essence, \autoref{eq:appx_modiste_uknn_dist} is the linear combination of distance between the given two contexts and distance between the two given actions, which accounts for similarity between contexts and actions at the same time.

\section{Experiment Details}\label{sec:appx_experimental}
\paragraph{Expert Queries} Each experiment used an initial (random) acquisition budget of $500$ expert queries and a total budget of $8,000$ queries.

\paragraph{Reward Estimators} The XGBoost ensemble score estimators consist of 2 XGBoost regressors (both trained on the entire set of collected examples) with \texttt{n\_estimators} set to 64. Our structured mimic score estimator uses ensembles consisting of XGBoost classifiers with \texttt{n\_estimators=64} for both the expert mimic and the ground truth classifier components of the model. For completeness, we also included Gaussian Process based score estimator with constant mean and radial basis function kernel; the parameters for mean and covariance functions are updated using Adam optimizer with learning rate set to $10^{-3}$; the GP update subroutine is called after each query.

\paragraph{Action Space} We limited the search space of possible actions for a given context to a maximum of $5000$ options; for any setting where the full size of the action space $\| \{0,1\}^d \times \mathbb{O} \| > 5000$, we searched over a random sample of $5000$ actions.

\begin{figure}[h]
    \centering
    \includegraphics[width=.7\linewidth]{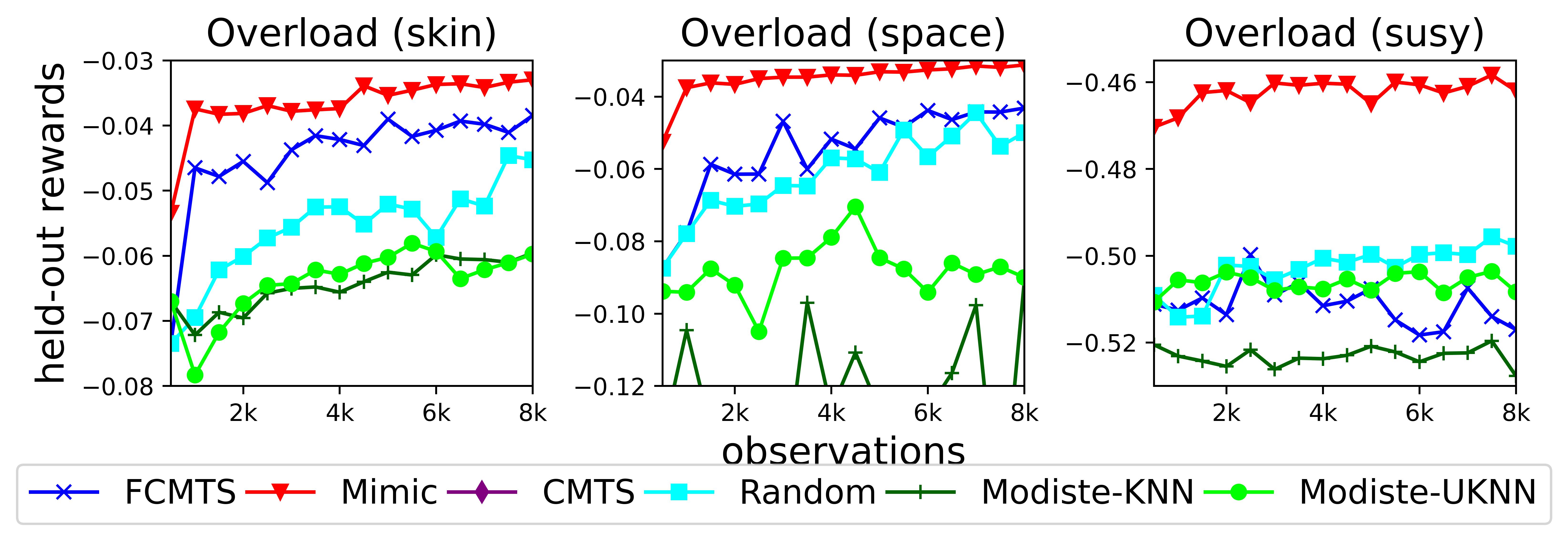}
    \caption{Average rewards vs.~observation data budget in the overload bias experimental setting, including additional baseline results for CMTS in `skin' (using a Gaussian Process, not able to achieve rewards in range).}
    \label{fig:overload_gp}
\end{figure}

\begin{figure}[h]
    \centering
    \includegraphics[width=.7\linewidth]{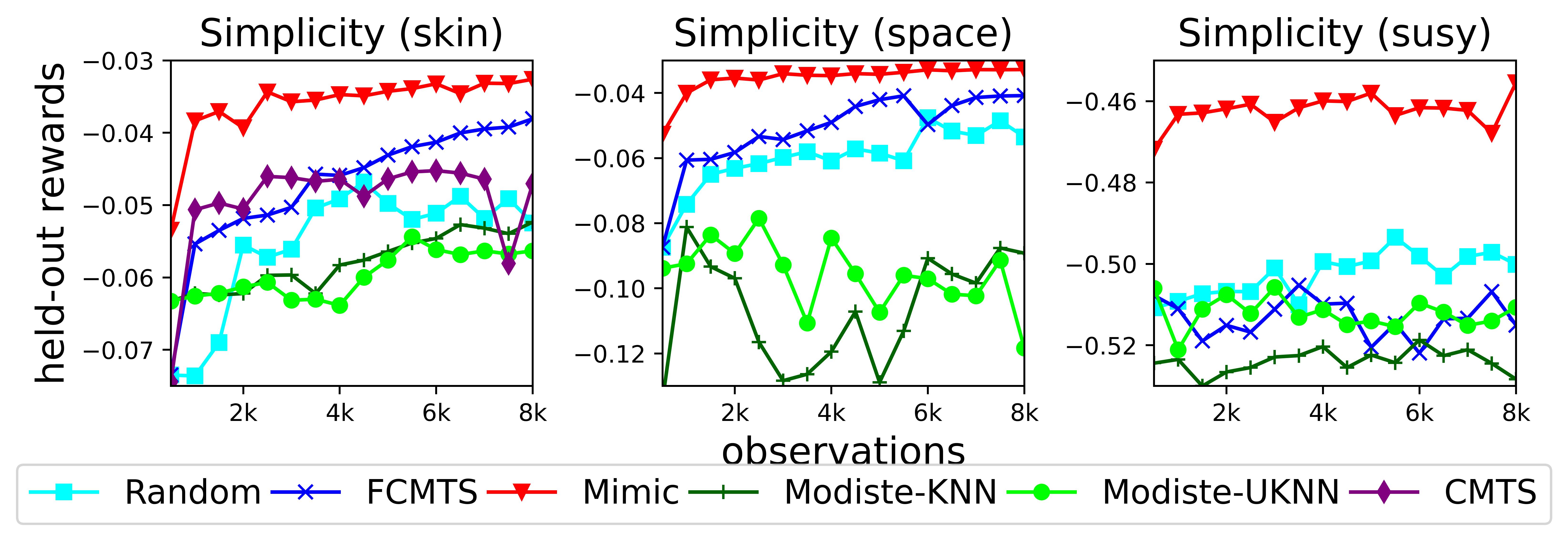}
    \caption{Average rewards vs.~observation data budget in the simplicity bias experimental setting, including additional baseline results for CMTS in `skin' (using a Gaussian Process).}
    \label{fig:poison_gp}
\end{figure}

\paragraph{Additional Discussion} We note that \texttt{CMTS} (CMTS \citep{Char2019} with GP score estimator) at times perform on par with \texttt{FCMTS} on smaller dimensional data (`skin'). However, on the \text{skin}-interpretability environment, \texttt{CMTS} takes 2.449 days to perform 8000 queries whereas our proposed \texttt{Mimic} only takes 4.039 hours; in addition, \texttt{CMTS} consumes significantly more memory compare to \texttt{Mimic}, which made it infeasible in even moderate dimensionalities notwithstanding GP optimizations. Furthermore, results below, and other preliminary experiments, show instability when optimizing GP hyperperameters (e.g. \autoref{fig:overload_gp},\autoref{fig:poison_gp}, \autoref{fig:beta_sweep_curves_appx_00}); in constrast, the \texttt{Mimic} approach was able to use consistent hyperparameters across datasets.

\subsection{Biased Synthetic Expert Setting}\label{sec:appx_experimental_synthetic_expert}
\paragraph{Overload Bias Experiments} The expert's \texttt{bias\_level} parameter was set to \texttt{0.5}, with \texttt{min\_temp=1.0} and \texttt{bias\_mult=5.0}.

\paragraph{Risk Aversion Bias Experiments} We set \texttt{bias\_level=0.5}.

\begin{figure}[h]
    \centering
    \includegraphics[width=.7\linewidth]{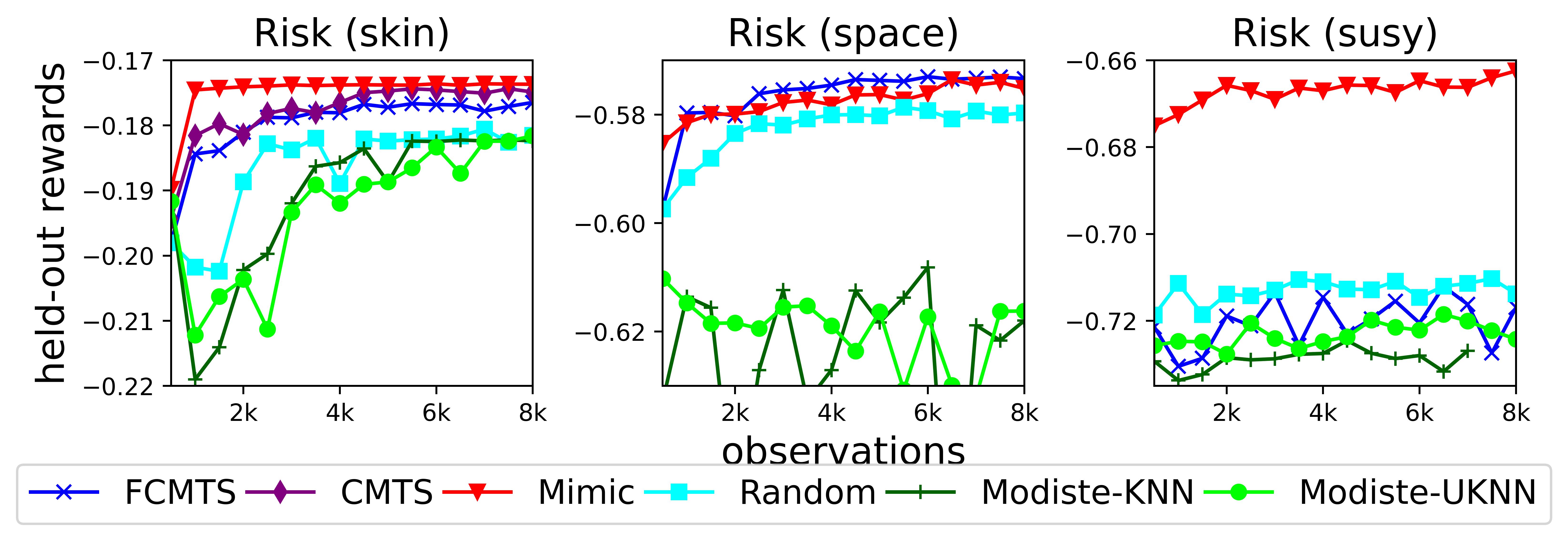}
    \caption{Average rewards vs.~observation data budget in the risk aversion bias experimental setting, including additional baseline results for CMTS  in `skin' (using a Gaussian Process).}
    \label{fig:risk_gp}
\end{figure}

\paragraph{Simplicity Bias Experiments} We set \texttt{bias\_level=0.5}; the \texttt{poison\_feature\_index} was set independently for each dataset by training univariate classifiers for the ground truth data on each feature separately and selecting the feature that resulted in the worst performance. This exploration led us to select feature \texttt{1} for the \texttt{skin} data, feature \texttt{5} for \texttt{space\_shuttle}, and feature \texttt{16} for \texttt{susy}.

\subsection{Multi-Expert Setting}\label{sec:appx_experimental_multi_expert}

\paragraph{Expert Configuration} As previously mentioned, we used K-Means clustering to simulate knowledge bases for different experts; for our experiments, we considered \texttt{k=4} experts and did not apply any added biases or limitations to the experts' decisions.

\begin{figure}[h]
    \centering
    \includegraphics[width=.7\linewidth]{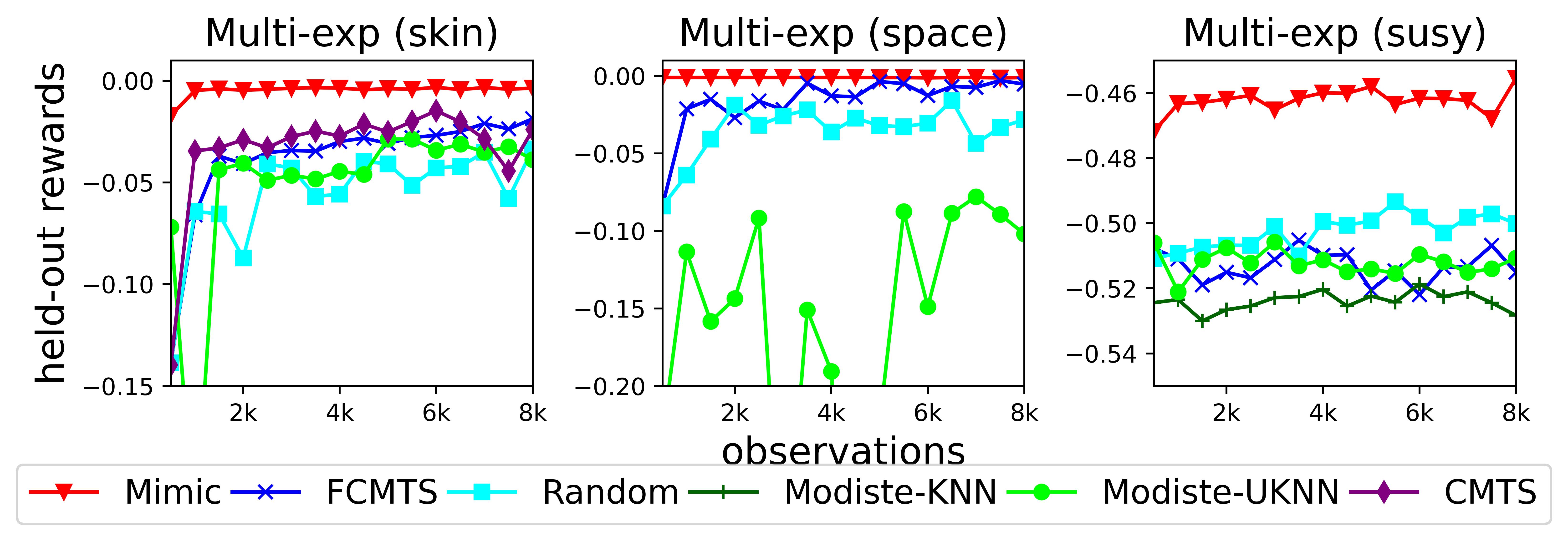}
    \caption{
        Average rewards vs.~observation data budget on the multi-expert environment including additional baseline results for CMTS using Gaussian Process in `skin.' (\texttt{Modiste-KNN} failed to produce rewards within reported ranges.)
    }
    \label{fig:multi-supp}
\end{figure}

\subsection{Visualization and Interpretability}\label{sec:appx_experimental_vis_and_interp}
\begin{figure}[h]
    \centering
    \begin{subfigure}{0.49\linewidth}
        \centering
        \includegraphics[width=\textwidth]{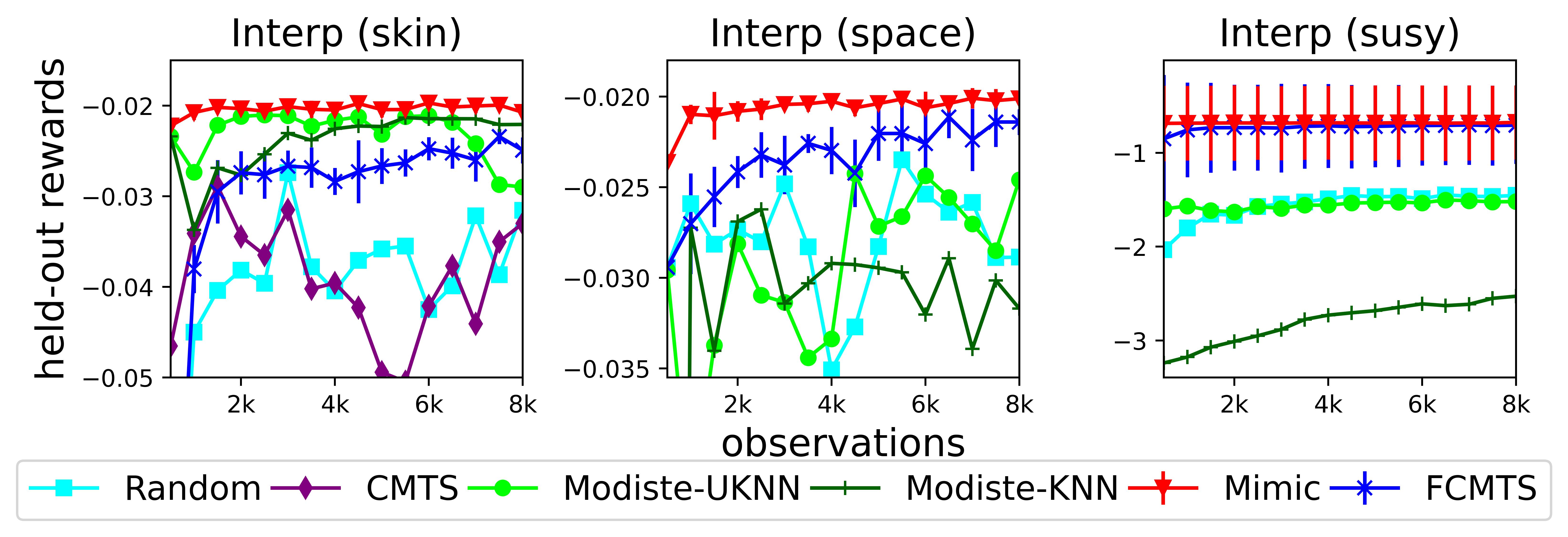}
        \caption{$\lambda=0.0$}
        \label{fig:beta_sweep_curves_appx_00}
    \end{subfigure}
    \begin{subfigure}{0.49\linewidth}
        \centering
        \includegraphics[width=\textwidth]{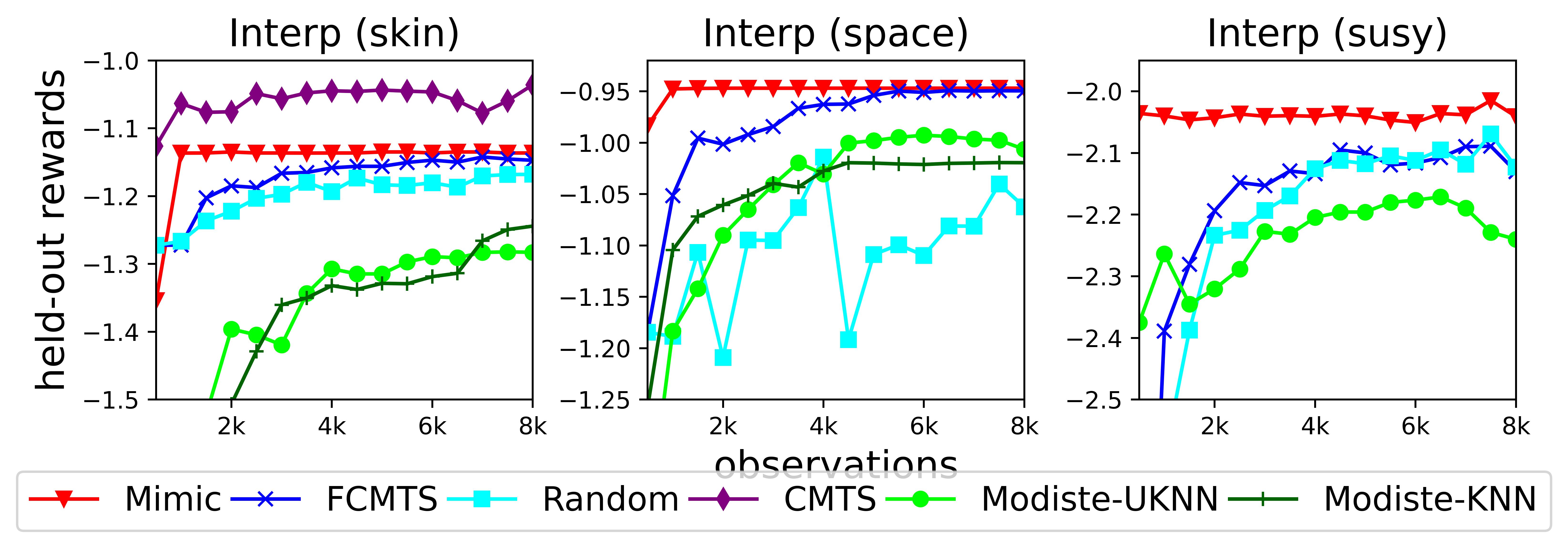}
        \caption{$\lambda=0.9$}
        \label{fig:beta_sweep_curves_appx_09}
    \end{subfigure}
    \caption{Average rewards vs.~observation data budget on the multi-expert environment. \texttt{Modiste-KNN} failed to produce rewards within reported ranges. \texttt{CMTS} is only run on skin segmentation dataset.}
    \label{fig:beta_sweep_curves_appx}
\end{figure}

We present the learning curve with complete set of baseline in \autoref{fig:beta_sweep_curves_appx}; among all experiments in \autoref{fig:beta_sweep_curves_appx}, \texttt{FCMTS} and \texttt{Mimic} are repeated six times and the error bar represents one sample standard deviation among the runs. From \autoref{fig:beta_sweep_curves_appx}, we observed that the learning curve is visually smoother as $\lambda$ increases; we conjecture that the strength of the penalty term ``eliminated'' actions that could reach the same negative cross-entropy loss but uses more features, which resulted in the policy has fewer options to consider during the learning phase.

\begin{figure}[h]
    \centering
    \begin{subfigure}{0.49\linewidth}
        \centering
        \includegraphics[width=\textwidth]{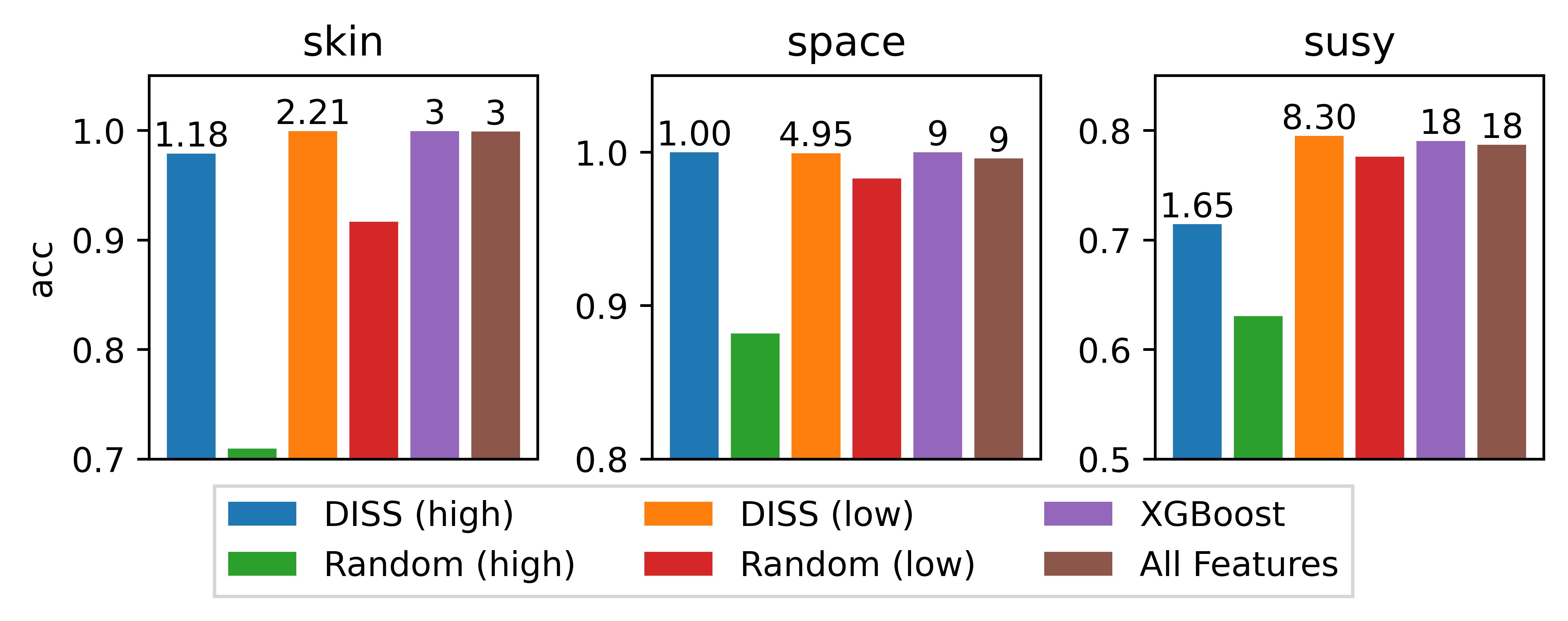}
        \caption{Accuracy}
        \label{fig:interp_auroc_0.5}
    \end{subfigure}
    \begin{subfigure}{0.49\linewidth}
        \centering
        \includegraphics[width=\textwidth]{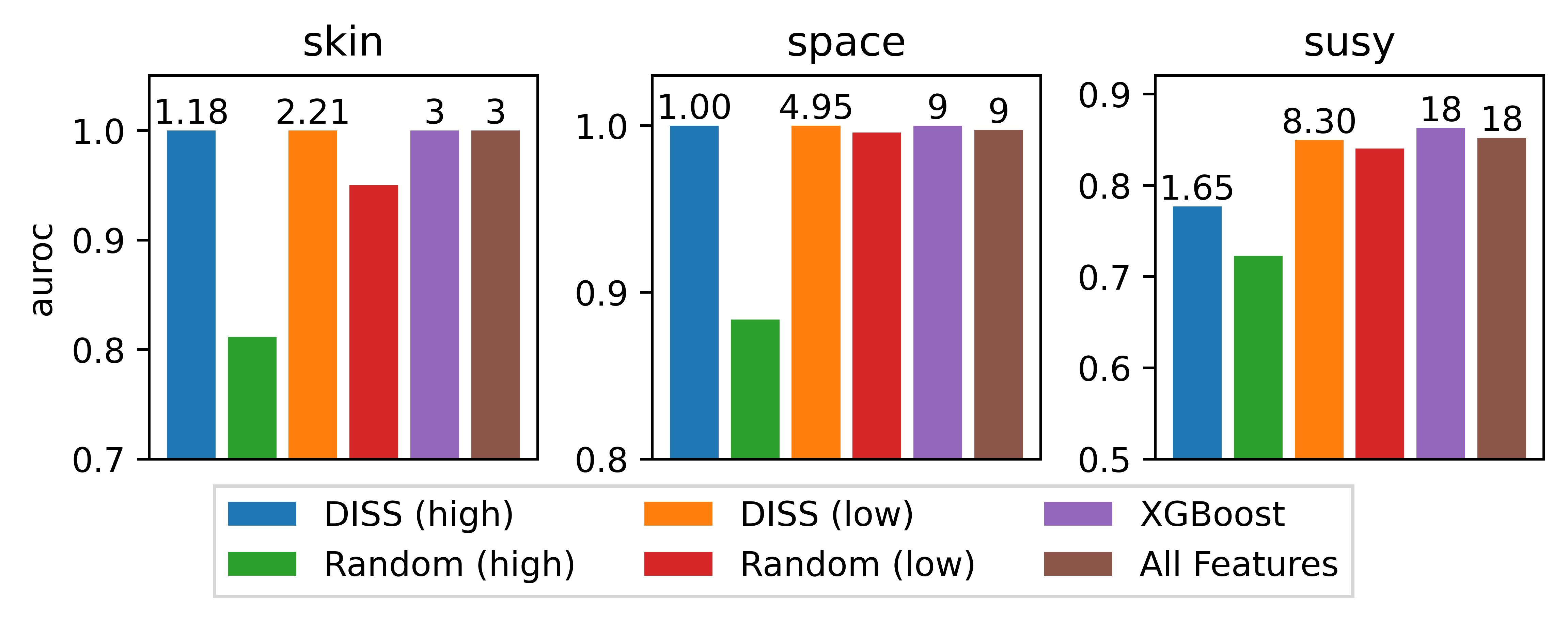}
        \caption{AUROC}
        \label{fig:interp_auroc_0.9}
    \end{subfigure}
    \caption{Accuracy and AUROC of \texttt{Mimic} selected features vs.~random features and full feature classification (average selected reported); low corresponds to $\lambda=0.0$ and high corresponds to $\lambda=0.9$. Numbers above the bar are the average number of features selected.}
    \label{fig:interp_bar}
\end{figure}

We also show Accuracy and AUROC of \texttt{Mimic} selected features vs.~random features and full feature classification (average selected reported) in \autoref{fig:interp_bar}. We note that number of features selected drastically decrease as lambda increases without degrading in task performance metrics. The phenomena are more significant in higher dimensional settings such as \texttt{space} and \texttt{susy}; on susy dataset, the policy on average select 8.30 features when $\lambda=0$ compare to 1.65 when $\lambda=0.9$; similar could also be observed for space dataset.

\begin{figure}[h]
    \centering
    \includegraphics[width=0.5\textwidth]{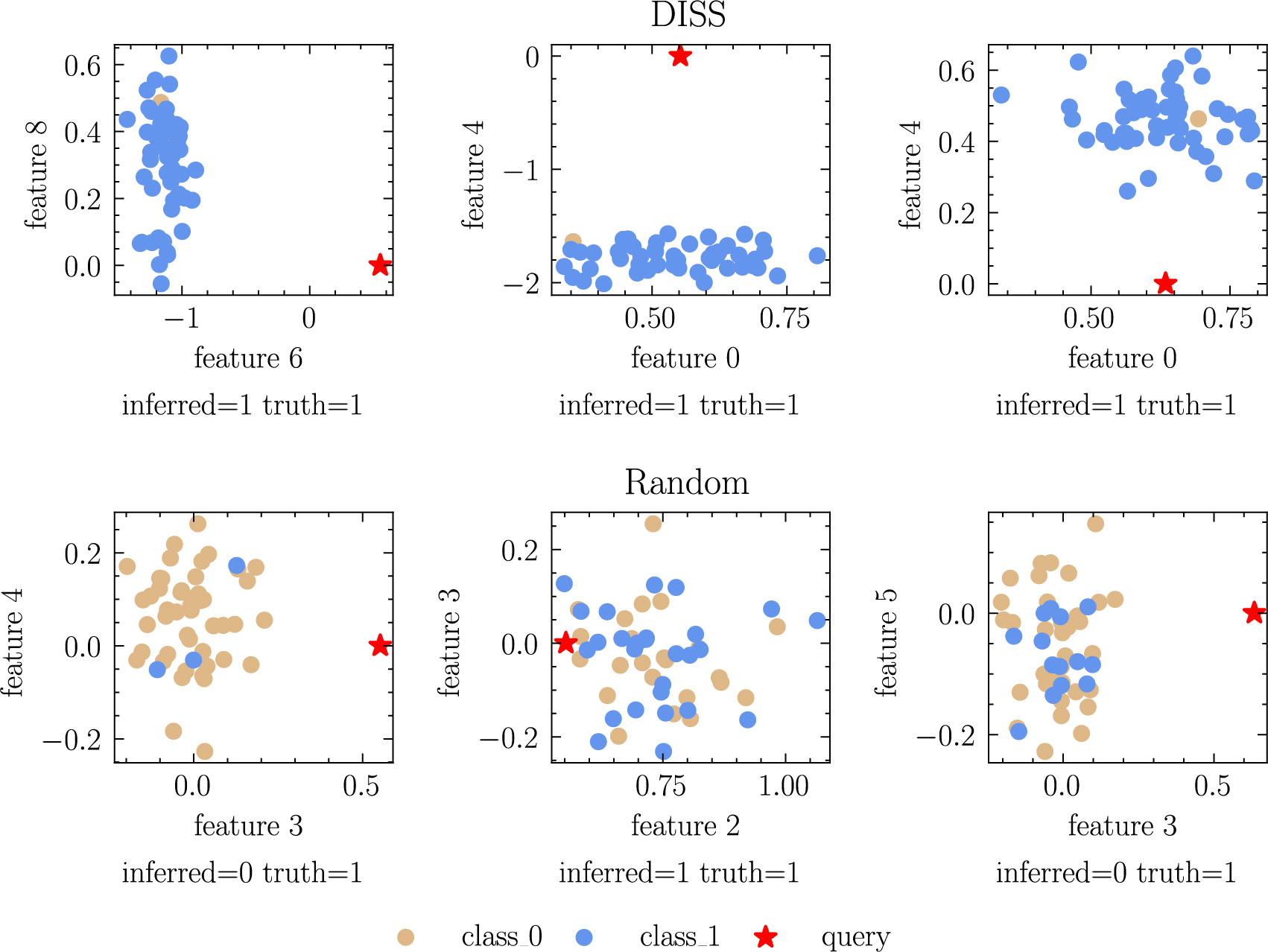}
    \caption{Two-dimensional visualization of the proposed KNNLRC model with subset of features selected with DISS policy versus that of chosen randomly.}
    \label{fig:appx_interp_model_display}
\end{figure}

In \autoref{fig:appx_interp_model_display}, we provide additional samples to demonstrate the capability of our proposed KNNLRC model in two-dimensional cases with space dataset. We plot the query instance in red star along with the nearest neighbors in the training set of the query with class label 0 and 1 in tan and blue dots, respectively. From the first column \autoref{fig:appx_interp_model_display}, the KNNLRC expert is able to make correct inference with high confidence (high label purity among neighbors of the query in the training set); however, if one were to randomly select two features, it is possible for the KNNLRC expert to make wrong inference while being extremely confident about its decision. From the second column, we can see that the KNNLRC expert is able to draw the correct conclusion but is extremely unconfident about its inference (low label purity among neighbors of the query in the training set), whereas features selected by our DISS policy is both correct and confident at its confident, which again demonstrate the quality of the features selected by DISS. In the last column, we show that it is also possible that randomly selected features result in KNNLRC expert to be less unconfident when making wrong decisions. From \autoref{fig:appx_interp_model_display}, we show features selected by DISS are informative and also our proposed KNNLRC interpretable model.

\subsection{Large Language Model Setting}\label{sec:appx_experimental_llm}
The following are examples contrasting the LLM outputs obtained using our \texttt{Mimic} approach vs \texttt{random} selection of features. In addition to the accuracy bar chart in the main text, we also provide an additional similar bar chart for AUROC in \autoref{fig:beta_sweep_curves_appx}. In the example for \texttt{Mimic} based subselection of features, notice the choice of salient features for the patient, enabling the LLM to generate better prediction, confidence and logical explanations. Compared to \texttt{random}, the LLM outputs fall short in prediction, confidence and explanations, leading to a misclassification(prediction of ``No'' diabetes for a true case). For all of the experiments, the factor that controls the randomness in output, \texttt{temperature}, was set to $0.7$. For each input case, we extract the features using policies described in previous sections which are incorporated in the prompt, guiding the LLM to make predictions based on the selected features. The LLM returns a natural text as an output that specifies prediction, probability, and an explanation for this prediction. We parse the text to select the probability of this prediction to return a reward.
\paragraph{Mimic based selection of features}
\begin{mdframed}
    \texttt{
    \item \textbf{Prompt:} You are an AI model trained to predict whether a patient has diabetes based on certain medical features. Your task is to analyze the given provided features and provide a prediction.
    \item \textbf{Some important notes:} Not all features may be provided for every prediction. Please interpret these as best as you can and state the explanation behind your prediction.
    \item You will be given the following features for a patient: Gender of the patient, Age of the patient (in years), Whether the patient has hypertension (0 for No, 1 for Yes), Whether the patient has heart disease (0 for No, 1 for Yes), Patient's smoking history, Body Mass Index (weight in kg / (height in m$^2$)), HbA1c level in blood (glycated hemoglobin), Blood glucose level (mg/dL)
    \item Strictly follow this output format
    \item 1. \textbf{Prediction:} State whether the patient is likely to have diabetes "Positive" or "Negative", and mention how confident you are [Low, Moderate, High]
    \item 2. \textbf{Confidence:} Express your confidence as a probability strictly in this format the scores should range from 0 to 1.
    \item 3. \textbf{Provide explanation for your predictions:}
    Please provide your prediction for the following: {\color{purple}Patient has hypertension, Patient's smoking history is True, HbA1c level is 6.099999904632568, and the blood glucose level is 160.0}
    }
\end{mdframed}
\textbf{Here's the LLM output:}
\begin{mdframed}
    \texttt{
        1. \textbf{Prediction:} Positive, High
        2. \textbf{Confidence:} 0.95
        3. \textbf{Explanation:}
        \item The patient has hypertension (1.0), which is a known risk factor for diabetes.
        \item The patient has a smoking history (1.0), which is also a risk factor for diabetes.
        \item The patient's HbA1c level (6.099999904632568) is above the normal range (typically < 5.7\%), indicating potential glucose regulation issues.
        \item The patient's blood glucose level (160.0 mg/dL) is higher than the normal range (typically < 100 mg/dL), further indicating potential glucose regulation issues.
        \item Given these factors, the patient's likelihood of having diabetes is high. However, I would like to note that a crucial piece of information is missing - the patient's age. Age is an important factor in diabetes prediction, and without it, my confidence in the prediction is slightly lower than it could be. Nevertheless, based on the available data, I am confident that the patient is likely to have diabetes.
    }
\end{mdframed}
\paragraph{Random strategy of selecting features in same patient:}
\begin{mdframed}
    \texttt{
        \item \textbf{Prompt:} \textcolor{red}{Same Preamble as in the Mimic case}
        \item \textbf{Provide explanation for your predictions}
        Please provide your prediction for the following {\color{purple}HbA1c level is 6.099999904632568, and the blood glucose level is 160.0}
        \item Strictly follow this output format:
        \textcolor{blue}{Same output format as in the Mimic case}
    } \newline
\end{mdframed}
\textbf{Here's the LLM output:}
\begin{mdframed}
    \texttt{
        1. \textbf{Prediction:} Negative, High
        2. \textbf{Confidence:} 0.05
        3. \textbf{Explanation:}
        \item The given data includes HbA1c level and blood glucose level, which are crucial for predicting diabetes. The HbA1c level is 6.099999904632568, which is higher than the normal range (typically < 5.7). The blood glucose level is 160.0 mg/dL, which is higher than the normal range (typically < 100 mg/dL). These values indicate that the patient may be at risk for diabetes.
        \item However, the data is missing some crucial features such as age, hypertension, heart disease, and bmi, which can also influence the prediction. Without these features, the prediction is based on the available data and may not be as accurate.
        \item Considering the available data, the patient's HbA1c level and blood glucose level suggest a low risk for diabetes.
    }
\end{mdframed}

\section{Additional Figures for LLM Experiments}\label{sec:appx_llm_examples}
\begin{figure}[h]
    \centering
    \includegraphics[width=0.75\linewidth]{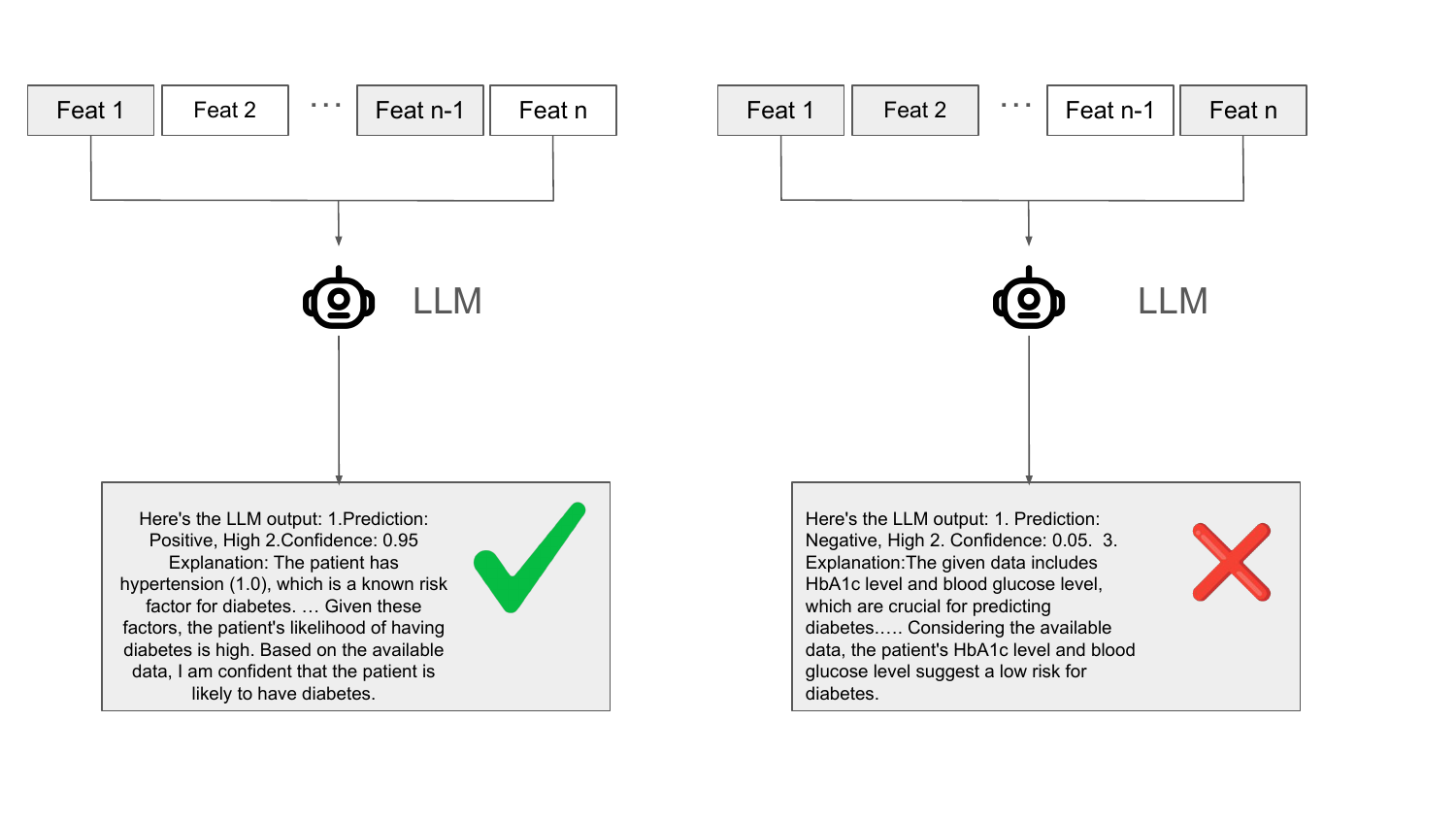}
    \caption{A graphical representation of our Mimic-based feature vs random feature selections(highlighted) sent to LLM as part of the prompt.}
    \label{fig:graphical_llm_abstract}
\end{figure}

Please see \autoref{fig:graphical_llm_abstract}.

\printbibliography[segment=1]

\end{document}